\documentclass[10pt,twocolumn,letterpaper]{article}
\usepackage{cvpr}
\usepackage{times}
\usepackage{epsfig}
\usepackage{graphicx}
\usepackage{amsmath}
\usepackage{amssymb}
\usepackage{enumitem}
\usepackage[nocompress]{cite}
\usepackage{xcolor}
\usepackage{multicol}
\usepackage{multirow}
\usepackage{makecell}
\usepackage{etoolbox}
\usepackage{caption}
\usepackage{adjustbox}
\usepackage{booktabs}
\usepackage{balance}
\usepackage{colortbl}
\usepackage{hhline}
\usepackage{lipsum}

\newcommand\blfootnote[1]{%
	\begingroup
	\renewcommand\thefootnote{}\footnote{#1}%
	\addtocounter{footnote}{-1}%
	\endgroup
}

\usepackage{tabularx}
\usepackage{booktabs}
\usepackage{float,wrapfig}
\usepackage{subcaption} 

\usepackage[pagebackref=true,breaklinks=true,letterpaper=true,colorlinks,bookmarks=false]{hyperref}

\usepackage{amsmath,amsfonts,bm}
\newlength{\mywidth}

\newcommand{\pixtopixhd}{pix2pixHD\xspace}
\newcommand{\pixtopixplusplus}{pix2pixHD++\xspace}
\newcommand{\sims}{SIMS\xspace}
\newcommand{\crn}{CRN\xspace}

\newcommand{\Ours}{Ours\xspace}

\newcommand{\mysubsection}[1]{\vspace{1mm}\noindent{\bf #1}}

\newcommand{\figsuppcoco}{\addtolength{\tabcolsep}{-4.5pt}    
\newlength{\mywidthsuppcoco}
\setlength{\mywidthsuppcoco}{0.193\textwidth}
\setlength{\fboxsep}{0pt}%
\setlength{\fboxrule}{0.5pt}%
\newcommand{\insertrow}[1]{
\fbox{\includegraphics[width=\mywidthsuppcoco, height=1.0in]{{"supplementary/coco/label/#1"}.jpg}} & 
\fbox{\includegraphics[width=\mywidthsuppcoco, height=1.0in]{{"supplementary/coco/gt/#1"}.jpg}} & \fbox{\includegraphics[width=\mywidthsuppcoco, height=1.0in]{{"supplementary/coco/crn/#1"}.jpg}} &
\fbox{\includegraphics[width=\mywidthsuppcoco, height=1.0in]{{"supplementary/coco/pix2pixhd/#1"}.jpg}} &
\fbox{\includegraphics[width=\mywidthsuppcoco, height=1.0in]{{"supplementary/coco/ours/#1"}.jpg}} \\
}
\bgroup
\def\arraystretch{1}
\begin{figure*}[t!]
\begin{tabular}{ccccc}
Label & Ground Truth & \crn & \pixtopixhd & \textbf{Ours} \\
\insertrow{000000134689}
\insertrow{000000193245}
\insertrow{000000024027}
\insertrow{000000051314}
\insertrow{000000445602}
\insertrow{000000407650}
\insertrow{000000436551}
\insertrow{000000383921}

\end{tabular}
	\caption{Additional results with comparison to those from the \crn~\cite{chen2017photographic} and \pixtopixhd~\cite{wang2018pix2pixHD} methods on the COCO-Stuff dataset.}\label{fig::supp_coco}
\end{figure*}
 \egroup
 \addtolength{\tabcolsep}{4.5pt}
\vfill

 }
\newcommand{\figsuppcocotwo}{\addtolength{\tabcolsep}{-4.5pt}    
\setlength{\mywidthsuppcoco}{0.193\textwidth}
\setlength{\fboxsep}{0pt}
\setlength{\fboxrule}{0.5pt}
\renewcommand{\insertrow}[1]{
\fbox{\includegraphics[width=\mywidthsuppcoco, height=1.0in]{{"supplementary/coco/label/#1"}.jpg}} & 
\fbox{\includegraphics[width=\mywidthsuppcoco, height=1.0in]{{"supplementary/coco/gt/#1"}.jpg}} & \fbox{\includegraphics[width=\mywidthsuppcoco, height=1.0in]{{"supplementary/coco/crn/#1"}.jpg}} &
\fbox{\includegraphics[width=\mywidthsuppcoco, height=1.0in]{{"supplementary/coco/pix2pixhd/#1"}.jpg}} &
\fbox{\includegraphics[width=\mywidthsuppcoco, height=1.0in]{{"supplementary/coco/ours/#1"}.jpg}} \\
}
\bgroup
\def\arraystretch{1}
\begin{figure*}[t!]
\begin{tabular}{ccccc}
Label & Ground Truth & \crn & \pixtopixhd & \textbf{Ours} \\
\insertrow{000000250758}
\insertrow{000000490470}
\insertrow{000000139684}
\insertrow{000000062025}
\insertrow{000000074733}
\insertrow{000000030785}
\insertrow{000000206994}
\insertrow{000000178744}
\end{tabular}
	\caption{Additional results with comparison to those from the \crn~\cite{chen2017photographic} and \pixtopixhd~\cite{wang2018pix2pixHD} methods on the COCO-Stuff dataset.} \label{fig::supp_coco2}	
	\end{figure*}
 \egroup
 \addtolength{\tabcolsep}{4.5pt}
\vfill

 }
\newcommand{\figsuppade}{\addtolength{\tabcolsep}{-4.5pt}    
\newlength{\mywidthsuppade}
\newlength{\myheightsuppade}
\setlength{\mywidthsuppade}{0.192\textwidth}
\setlength{\myheightsuppade}{1.1in}
\setlength{\fboxsep}{0pt}%
\setlength{\fboxrule}{0.5pt}%
\renewcommand{\insertrow}[1]{
\fbox{\includegraphics[width=\mywidthsuppade, height=\myheightsuppade]{{"supplementary/ade20k/label/ADE_val_#1"}.jpg}} & 
\fbox{\includegraphics[width=\mywidthsuppade, height=\myheightsuppade]{{"supplementary/ade20k/gt/ADE_val_#1"}.jpg}} & \fbox{\includegraphics[width=\mywidthsuppade, height=\myheightsuppade]{{"supplementary/ade20k/crn/ADE_val_#1"}.jpg}} &
\fbox{\includegraphics[width=\mywidthsuppade, height=\myheightsuppade]{{"supplementary/ade20k/pix2pixhd/ADE_val_#1"}.jpg}} &
\fbox{\includegraphics[width=\mywidthsuppade, height=\myheightsuppade]{{"supplementary/ade20k/ours/ADE_val_#1"}.jpg}} \\
}
\bgroup
\def\arraystretch{1}
\begin{figure*}[t!]
\begin{tabular}{ccccc}
Label & Ground Truth & CRN & \pixtopixhd & \textbf{Ours} \\
\insertrow{00000135}
\insertrow{00000251}
\insertrow{00000268}
\insertrow{00000309}
\insertrow{00000318}
\insertrow{00001012}
\insertrow{00001136}
\end{tabular}
	\caption{Additional results with comparison to those from the \crn~\cite{chen2017photographic} and \pixtopixhd~\cite{wang2018pix2pixHD} methods on the ADE20K dataset.} \label{fig::supp_ade}
 \end{figure*}
 \egroup
 \addtolength{\tabcolsep}{4.5pt}
 \vfill

 }
\newcommand{\figsuppadeoutdoor}{\addtolength{\tabcolsep}{-4.5pt}
\newlength{\mywidthsuppadeoutdoor}
\newlength{\myheightsuppadeoutdoor}
\setlength{\mywidthsuppadeoutdoor}{0.16\textwidth}
\setlength{\myheightsuppadeoutdoor}{0.73in}
\setlength{\fboxsep}{0pt}%
\setlength{\fboxrule}{0.5pt}%
\renewcommand{\insertrow}[1]{
\fbox{\includegraphics[width=\mywidthsuppadeoutdoor, height=\myheightsuppadeoutdoor]{{"supplementary/ade20k/label/ADE_val_#1"}.jpg}} &
\fbox{\includegraphics[width=\mywidthsuppadeoutdoor, height=\myheightsuppadeoutdoor]{{"supplementary/ade20k/gt/ADE_val_#1"}.jpg}} & \fbox{\includegraphics[width=\mywidthsuppadeoutdoor, height=\myheightsuppadeoutdoor]{{"supplementary/ade20k/crn/ADE_val_#1"}.jpg}} &
\fbox{\includegraphics[width=\mywidthsuppadeoutdoor, height=\myheightsuppadeoutdoor]{{"supplementary/ade20k/sims/ADE_val_#1"}.jpg}} &
\fbox{\includegraphics[width=\mywidthsuppadeoutdoor, height=\myheightsuppadeoutdoor]{{"supplementary/ade20k/pix2pixhd/ADE_val_#1"}.jpg}} &
\fbox{\includegraphics[width=\mywidthsuppadeoutdoor, height=\myheightsuppadeoutdoor]{{"supplementary/ade20k/ours/ADE_val_#1"}.jpg}} \\
}
\bgroup
\def\arraystretch{1}
\begin{figure*}[t!]
\begin{tabular}{cccccc}
Label & Ground Truth & \crn & \sims & \pixtopixhd & \textbf{Ours} \\
\insertrow{00000262}
\insertrow{00000289}
\insertrow{00000291}
\insertrow{00000389}
\insertrow{00000574}
\insertrow{00000733}
\insertrow{00001112}
\insertrow{00001344}
\insertrow{00001390}
\insertrow{00001582}
\insertrow{00001795}
\end{tabular}
	\caption{Additional results with comparison to those from the \crn~\cite{chen2017photographic}, \sims~\cite{qi2018semi}, and \pixtopixhd~\cite{wang2018pix2pixHD} methods on the ADE20K-outdoor dataset.} \label{fig::supp_adeoutdoor}		
 \end{figure*}
 \egroup
 \addtolength{\tabcolsep}{4.5pt}
\vfill

 }
\newcommand{\figsuppcityscape}{\addtolength{\tabcolsep}{-4.5pt}    
\newlength{\mywidthsuppcityscapes}
\setlength{\mywidthsuppcityscapes}{0.327\textwidth}
\setlength{\fboxsep}{0pt}%
\setlength{\fboxrule}{0.5pt}%
\renewcommand{\insertrow}[1]{
Label & Ground Truth & \textbf{Ours} \\ 
\fbox{\includegraphics[width=\mywidthsuppcityscapes]{{"supplementary/cityscapes/label/#1_gtFine_color"}.jpg}} & 
\fbox{\includegraphics[width=\mywidthsuppcityscapes]{{"supplementary/cityscapes/gt/#1_leftImg8bit"}.jpg}} & \fbox{\includegraphics[width=\mywidthsuppcityscapes]{{"supplementary/cityscapes/ours/#1_leftImg8bit"}.jpg}} \\
\fbox{\includegraphics[width=\mywidthsuppcityscapes]{{"supplementary/cityscapes/crn/#1_leftImg8bit"}.jpg}} &
\fbox{\includegraphics[width=\mywidthsuppcityscapes]{{"supplementary/cityscapes/sims/#1_leftImg8bit"}.jpg}} &
\fbox{\includegraphics[width=\mywidthsuppcityscapes]{{"supplementary/cityscapes/pix2pixhd/#1_leftImg8bit"}.jpg}} \\
CRN & SIMS & \pixtopixhd \\
}
\bgroup
\def\arraystretch{0.5}
\begin{figure*}[t!]
\begin{tabular}{ccc}
\insertrow{lindau_000004_000019}
\addlinespace[16pt]
\insertrow{munster_000013_000019}
\addlinespace[16pt]
\insertrow{munster_000096_000019}
\end{tabular}
	\caption{Additional results with comparison to those from the \crn~\cite{chen2017photographic}, \sims~\cite{qi2018semi}, and \pixtopixhd~\cite{wang2018pix2pixHD} methods on the Cityscapes dataset.} \label{fig::supp_cityscapes}		
 \end{figure*}
 \egroup
 \addtolength{\tabcolsep}{4.5pt}
\vfill

 }
\newcommand{\figsuppflickr}{\addtolength{\tabcolsep}{-4.5pt}    
\newlength{\mywidthsuppflickr}
\setlength{\mywidthsuppflickr}{0.19\textwidth}
\setlength{\fboxsep}{0pt}%
\setlength{\fboxrule}{0.5pt}%
\bgroup
\def\arraystretch{1}
\begin{figure*}[t!]
	\begin{tabular}{cc|ccc}
		Label & Ground Truth & & Multi-modal results & \\
		\fbox{\includegraphics[width=\mywidthsuppflickr, height=1.0in]{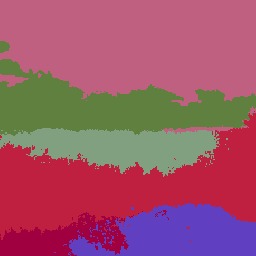}} & 
		\fbox{\includegraphics[width=\mywidthsuppflickr, height=1.0in]{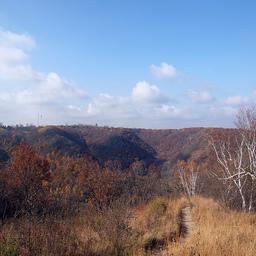}} & \fbox{\includegraphics[width=\mywidthsuppflickr, height=1.0in]{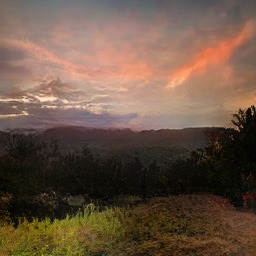}} &
		\fbox{\includegraphics[width=\mywidthsuppflickr, height=1.0in]{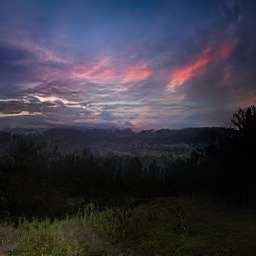}} &
		\fbox{\includegraphics[width=\mywidthsuppflickr, height=1.0in]{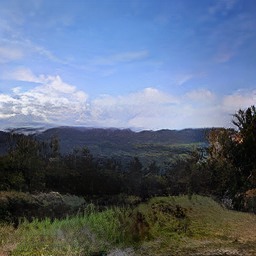}} \\ 
		
		\fbox{\includegraphics[width=\mywidthsuppflickr, height=1.0in]{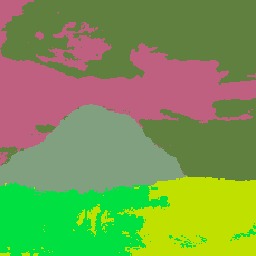}} & 
		\fbox{\includegraphics[width=\mywidthsuppflickr, height=1.0in]{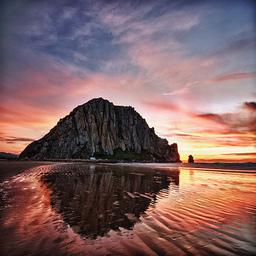}} & \fbox{\includegraphics[width=\mywidthsuppflickr, height=1.0in]{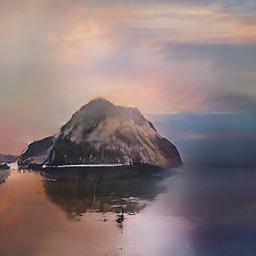}} &
		\fbox{\includegraphics[width=\mywidthsuppflickr, height=1.0in]{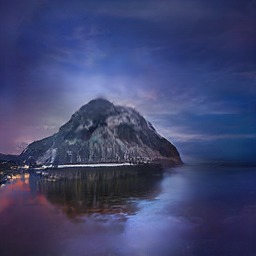}} &
		\fbox{\includegraphics[width=\mywidthsuppflickr, height=1.0in]{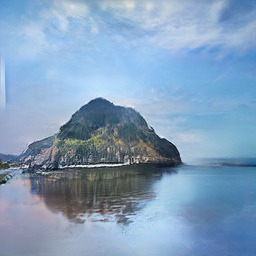}} \\
		
		\fbox{\includegraphics[width=\mywidthsuppflickr, height=1.0in]{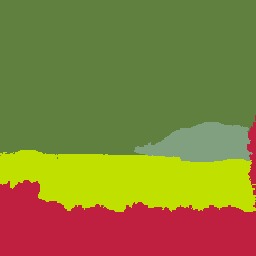}} & 
		\fbox{\includegraphics[width=\mywidthsuppflickr, height=1.0in]{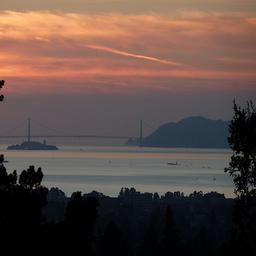}} & \fbox{\includegraphics[width=\mywidthsuppflickr, height=1.0in]{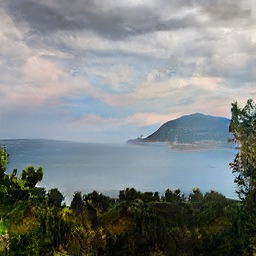}} &
		\fbox{\includegraphics[width=\mywidthsuppflickr, height=1.0in]{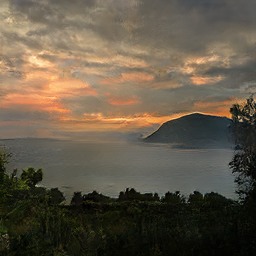}} &
		\fbox{\includegraphics[width=\mywidthsuppflickr, height=1.0in]{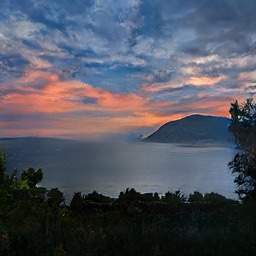}} \\
		
		\fbox{\includegraphics[width=\mywidthsuppflickr, height=1.0in]{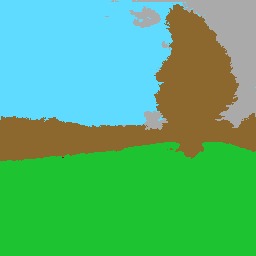}} & 
		\fbox{\includegraphics[width=\mywidthsuppflickr, height=1.0in]{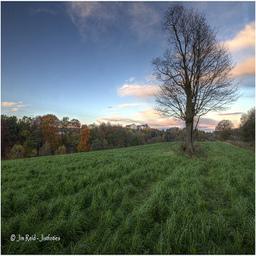}} & \fbox{\includegraphics[width=\mywidthsuppflickr, height=1.0in]{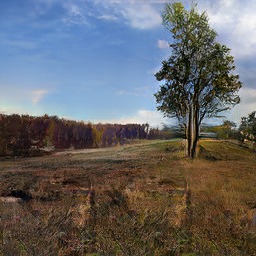}} &
		\fbox{\includegraphics[width=\mywidthsuppflickr, height=1.0in]{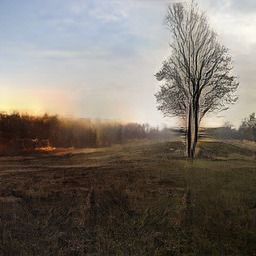}} &
		\fbox{\includegraphics[width=\mywidthsuppflickr, height=1.0in]{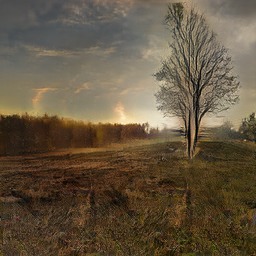}} \\
		
		\fbox{\includegraphics[width=\mywidthsuppflickr, height=1.0in]{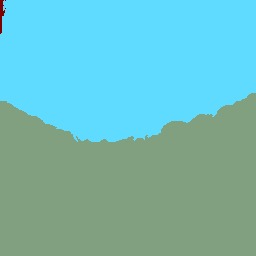}} & 
		\fbox{\includegraphics[width=\mywidthsuppflickr, height=1.0in]{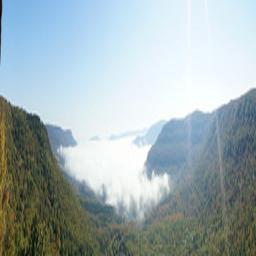}} & \fbox{\includegraphics[width=\mywidthsuppflickr, height=1.0in]{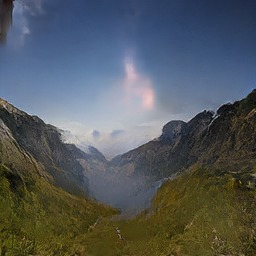}} &
		\fbox{\includegraphics[width=\mywidthsuppflickr, height=1.0in]{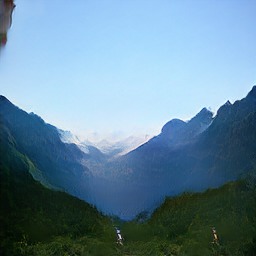}} &
		\fbox{\includegraphics[width=\mywidthsuppflickr, height=1.0in]{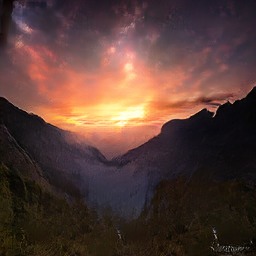}} \\
		
		\fbox{\includegraphics[width=\mywidthsuppflickr, height=1.0in]{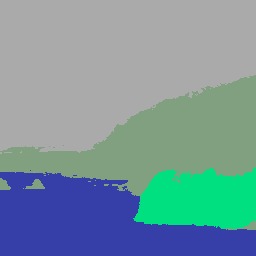}} & 
		\fbox{\includegraphics[width=\mywidthsuppflickr, height=1.0in]{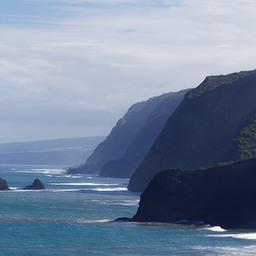}} & \fbox{\includegraphics[width=\mywidthsuppflickr, height=1.0in]{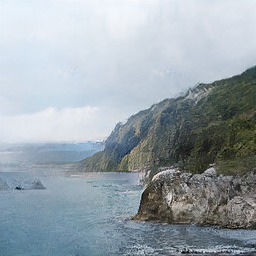}} &
		\fbox{\includegraphics[width=\mywidthsuppflickr, height=1.0in]{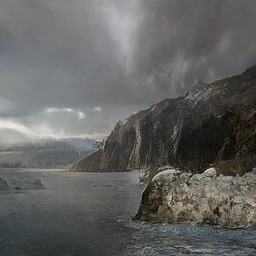}} &
		\fbox{\includegraphics[width=\mywidthsuppflickr, height=1.0in]{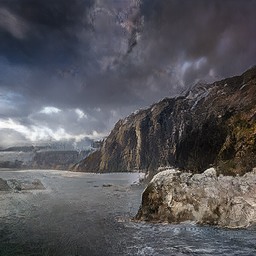}} \\
		
		\fbox{\includegraphics[width=\mywidthsuppflickr, height=1.0in]{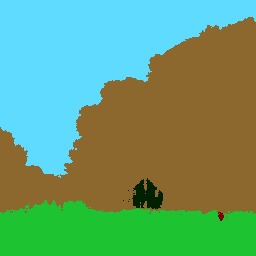}} & 
		\fbox{\includegraphics[width=\mywidthsuppflickr, height=1.0in]{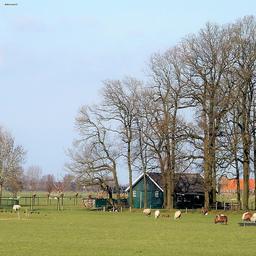}} & \fbox{\includegraphics[width=\mywidthsuppflickr, height=1.0in]{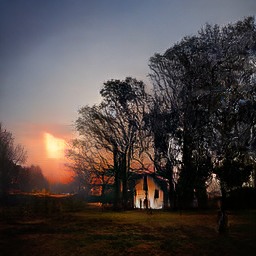}} &
		\fbox{\includegraphics[width=\mywidthsuppflickr, height=1.0in]{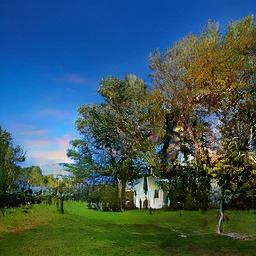}} &
		\fbox{\includegraphics[width=\mywidthsuppflickr, height=1.0in]{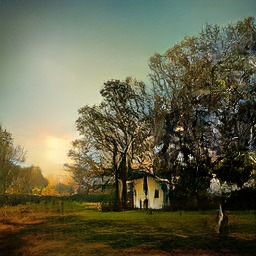}} \\
		
		\fbox{\includegraphics[width=\mywidthsuppflickr, height=1.0in]{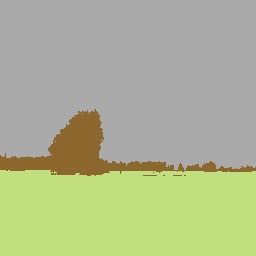}} & 
		\fbox{\includegraphics[width=\mywidthsuppflickr, height=1.0in]{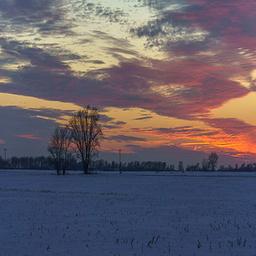}} & \fbox{\includegraphics[width=\mywidthsuppflickr, height=1.0in]{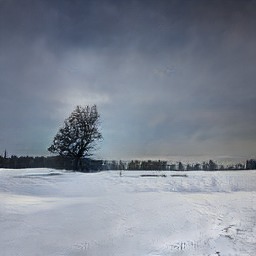}} &
		\fbox{\includegraphics[width=\mywidthsuppflickr, height=1.0in]{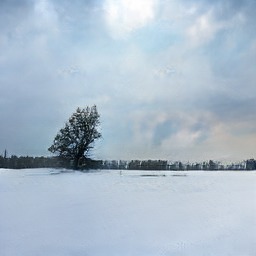}} &
		\fbox{\includegraphics[width=\mywidthsuppflickr, height=1.0in]{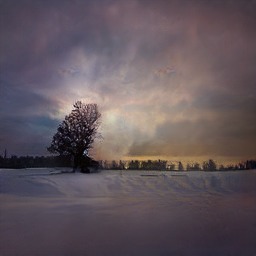}} \\
	\end{tabular}
	\caption{Additional multi-modal synthesis results on the Flickr Landscapes Dataset. By sampling latent vectors from a standard Gaussian distribution, we synthesize images of diverse appearances.} \label{fig::supp_flickr}		
\end{figure*}
\egroup
\addtolength{\tabcolsep}{4.5pt}
\vfill

}

\newcommand{\tblsuppablation}{\bgroup
\def\arraystretch{1.1}
\addtolength{\tabcolsep}{-2.5pt}  
\begin{table}[!h]
	\centering
	\small
    \begin{tabular}{l||c|c|c}
	    Method & COCO. & ADE. & City.\\
		\hline
Ours & 35.2 & 38.5 & 62.3 \\
Ours w/o  Perceptual loss & 24.7 & 30.1 & 57.4\\   
Ours w/o GAN feature matching loss & 33.2 & 38.0 & 62.2\\   
Ours w/\ \ \ a deeper discriminator & 34.9 & 38.3 & 60.9\\
\hline
\pixtopixplusplus w/\ \ \  SPADE & 
                34.4 & 39.0 & 62.2 \\
\pixtopixplusplus &
                32.7 & 38.3 & 58.8\\
\pixtopixplusplus w/o Sync BatchNorm & 
                27.4 & 31.8 & 51.1\\
\pixtopixplusplus w/o Sync BatchNorm, & 
                26.0 & 31.9 & 52.3\\
\quad \quad \quad \medspace and w/o Spectral Norm & & &\\
\hline
\pixtopixhd ~\cite{wang2018pix2pixHD}& 14.6 & 20.3 & 58.3\\
	\end{tabular}
	\caption{Additional ablation study results using the mIoU metric: the table shows that both the perceptual loss and GAN feature matching loss terms are important. Making the discriminator deeper does not lead to a performance boost. The table also shows that all the components (Synchronized BatchNorm, Spectral Norm, TTUR, the Hinge loss objective, and the SPADE) used in the proposed method helps our strong baseline, \pixtopixplusplus.}
	\label{tbl::moreablation}		
 \end{table}
 \addtolength{\tabcolsep}{2.5pt}  
 \egroup
 }

\newcommand{\tbleffectiveness}{\bgroup
\def\arraystretch{1.0}
\addtolength{\tabcolsep}{-2.5pt}  
\begin{table}[!t]
	\centering
	\small
    \begin{tabular}{r||c||c|c|c}
		Method & \#param & COCO. & ADE. & City.  \\
		\Xhline{2\arrayrulewidth}
		{\bf decoder w/ SPADE (\Ours)}  &	96M	& \textbf{35.2}	& 38.5	& 62.3 \\
		{\bf compact decoder w/ SPADE }  &	61M	& \textbf{35.2} & 38.0 & \textbf{62.5} \\
        decoder w/ Concat  &	79M	& 31.9	& 33.6	& 61.1 \\
        \hline
        {\bf \pixtopixplusplus w/ SPADE}	& 237M	&34.4	&\textbf{39.0}	& 62.2 \\
        \pixtopixplusplus w/ Concat	& 195M	& 32.9	& 38.9	& 57.1 \\
        \pixtopixplusplus 	& 183M	& 32.7 &	38.3	&58.8 \\
        compact \pixtopixplusplus & 103M &	31.6 & 37.3	& 57.6 \\
        \pixtopixhd ~\cite{wang2018pix2pixHD}	& 183M	& 14.6	& 20.3	& 58.3 \\
	\end{tabular}\vspace{-2mm}
	\caption{The mIoU scores are boosted when the SPADE is used, for both the decoder architecture  (Figure \ref{fig::architecture}) and encoder-decoder architecture of \pixtopixplusplus (our improved baseline over \pixtopixhd~\cite{wang2018pix2pixHD}). On the other hand, simply concatenating semantic input at every layer fails to do so. Moreover, our compact model with smaller depth at all layers outperforms all the baselines.}
	\label{tbl::effectiveness}		
 \vspace{-3mm}
 \end{table}
 \addtolength{\tabcolsep}{2.5pt}  
 \egroup
 }

\newcommand{\figcoco}{\addtolength{\tabcolsep}{-5pt}    

\newlength{\mywidthcoco}
\setlength{\mywidthcoco}{0.193\textwidth}
\newlength{\myheightcoco}

\bgroup
\def\arraystretch{0.5}
\begin{figure*}[]
\begin{tabular}{ccccc}
\setlength{\myheightcoco}{1.1in}
\setlength{\myheightcoco}{1.1in}
 \includegraphics[width=\mywidthcoco, height=1.1in]{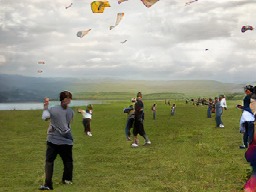} & \includegraphics[width=\mywidthcoco, height=1.1in]{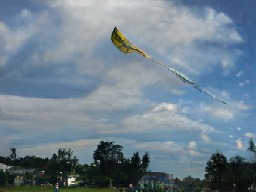} & \includegraphics[width=\mywidthcoco, height=1.1in]{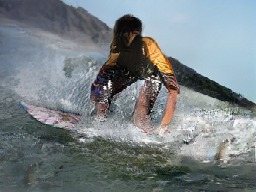} & \includegraphics[width=\mywidthcoco, height=1.1in]{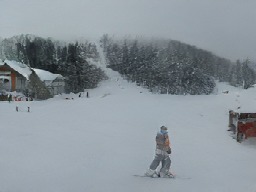} & \includegraphics[width=\mywidthcoco, height=1.1in]{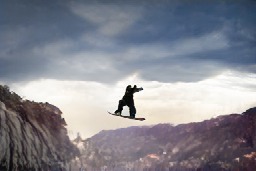} \\
  \setlength{\myheightcoco}{1.1in}
 \includegraphics[width=\mywidthcoco,height=1.1in]{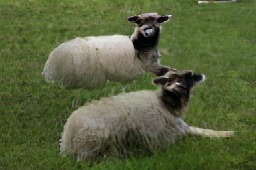} &
 \includegraphics[width=\mywidthcoco,height=1.1in]{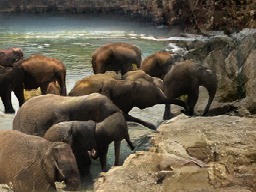} &
 \includegraphics[width=\mywidthcoco,height=1.1in]{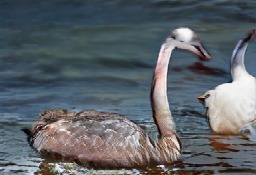} &
 \includegraphics[width=\mywidthcoco,height=1.1in]{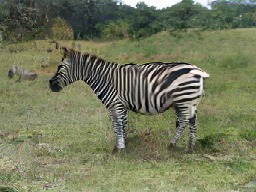} &
 \includegraphics[width=\mywidthcoco,height=1.1in]{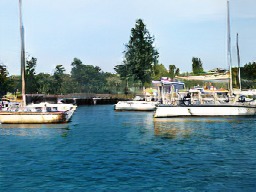} \\
 \includegraphics[width=\mywidthcoco,height=1.2in]{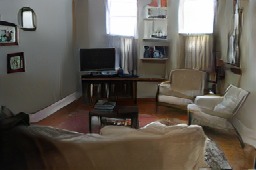} &
 \includegraphics[width=\mywidthcoco,height=1.2in]{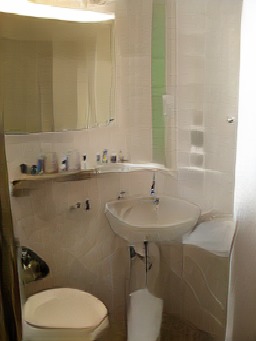} &
 \includegraphics[width=\mywidthcoco,height=1.2in]{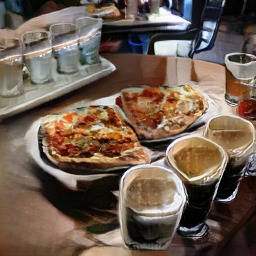} &
 \includegraphics[width=\mywidthcoco,height=1.2in]{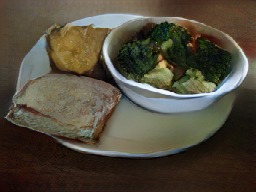} &
 \includegraphics[width=\mywidthcoco,height=1.2in]{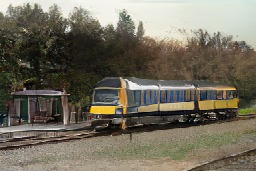} \\
\end{tabular}
\vspace{-3mm}
	\caption{Semantic image synthesis results on COCO-Stuff. Our method successfully generates realistic images in diverse scenes ranging from animals to sports activities.} \label{fig::coco}		
  	\vspace{-4mm}
 \end{figure*}
 \egroup
 \addtolength{\tabcolsep}{5pt}
}
\newcommand{\figcococomparison}{\addtolength{\tabcolsep}{-4.5pt}    
\setlength{\mywidth}{0.1932\textwidth}
\bgroup
\def\arraystretch{0.5}
\begin{figure*}[]
\begin{tabular} {cc|cc|c}
 Label & Ground Truth & CRN~\cite{chen2017photographic} &  \pixtopixhd~\cite{wang2018pix2pixHD} & Ours \\
 \includegraphics[width=\mywidth, height=0.79in]{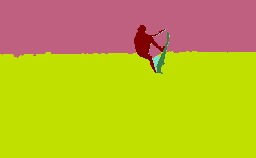} & \includegraphics[width=\mywidth, height=0.79in]{images/coco/gt/000000263403.jpg} &
 \includegraphics[width=\mywidth, height=0.79in]{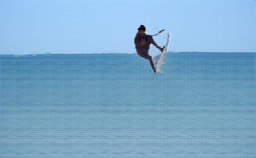} &   \includegraphics[width=\mywidth, height=0.79in]{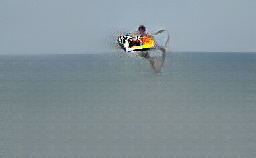} &  \includegraphics[width=\mywidth, height=0.79in]{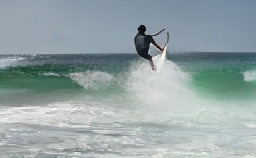} \\
 \includegraphics[width=\mywidth, height=0.79in]{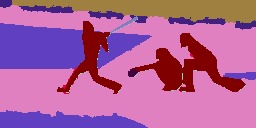} & \includegraphics[width=\mywidth, height=0.79in]{images/coco/gt/000000129945.jpg} &
 \includegraphics[width=\mywidth, height=0.79in]{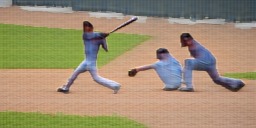} &   \includegraphics[width=\mywidth, height=0.79in]{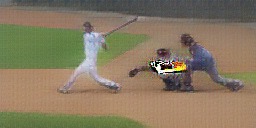} &  \includegraphics[width=\mywidth, height=0.79in]{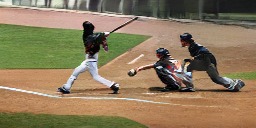} \\
\end{tabular}
\vspace{-2mm}
	\caption{Visual comparison of semantic image synthesis results on the COCO-Stuff dataset. Our method successfully synthesizes realistic details from semantic labels. }
	\label{fig::cococomparison}	
\vspace{-4mm}	
 \end{figure*}
 \egroup
 \addtolength{\tabcolsep}{4.5pt}
 
}
\newcommand{\figadecomparison}{\addtolength{\tabcolsep}{-4.5pt}    

\setlength{\mywidth}{0.16\textwidth}

\bgroup
\def\arraystretch{0.5}
\begin{figure*}[]
\begin{tabular} {cc|ccc|c}
 Label & Ground Truth & CRN~\cite{chen2017photographic} & SIMS~\cite{qi2018semi} & \pixtopixhd~\cite{wang2018pix2pixHD} & Ours \\
 
\includegraphics[width=\mywidth, height=0.80in]{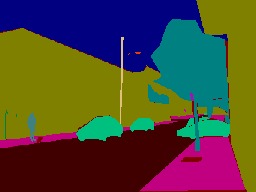} & \includegraphics[width=\mywidth, height=0.80in]{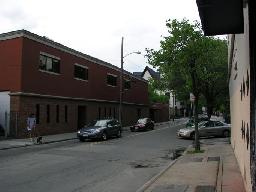} &
\includegraphics[width=\mywidth, height=0.80in]{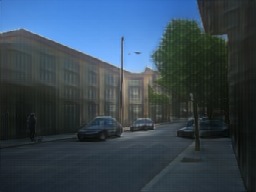} &   \includegraphics[width=\mywidth, height=0.80in]{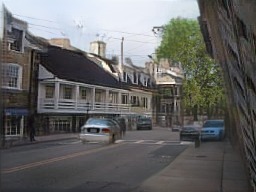} &
\includegraphics[width=\mywidth, height=0.80in]{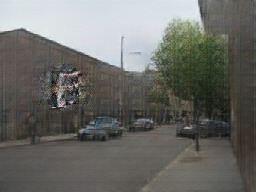} &  \includegraphics[width=\mywidth, height=0.80in]{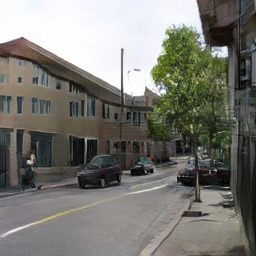} \\

\includegraphics[width=\mywidth, height=0.80in]{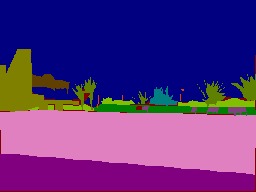} & \includegraphics[width=\mywidth, height=0.80in]{images/ade20k/gt/ADE_val_00001880.jpg} &
\includegraphics[width=\mywidth, height=0.80in]{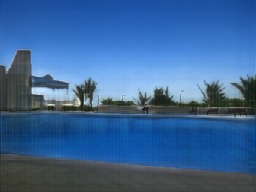} &  \includegraphics[width=\mywidth, height=0.80in]{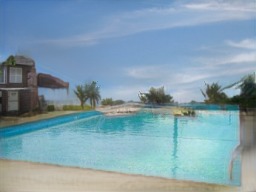} &
\includegraphics[width=\mywidth, height=0.80in]{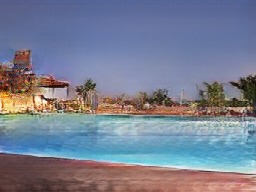} &  \includegraphics[width=\mywidth, height=0.80in]{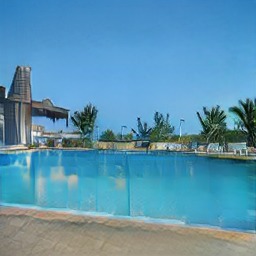} \\

\includegraphics[width=\mywidth]{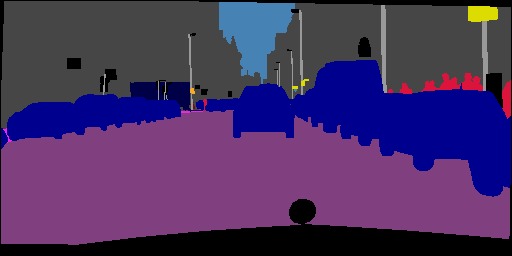} & 
\includegraphics[width=\mywidth]{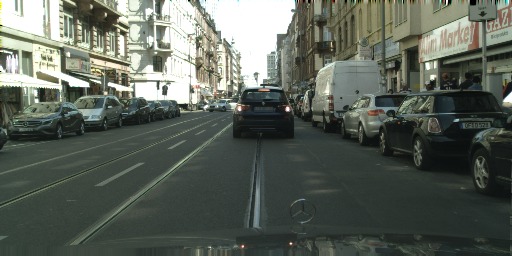} &
\includegraphics[width=\mywidth]{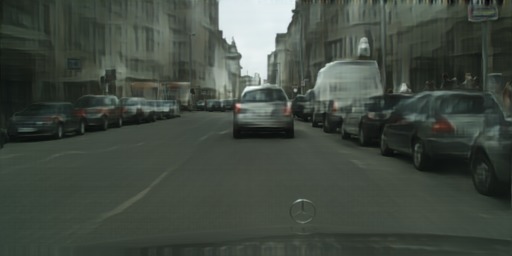} &
\includegraphics[width=\mywidth]{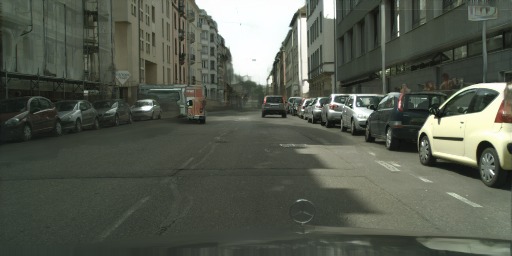} & 
\includegraphics[width=\mywidth]{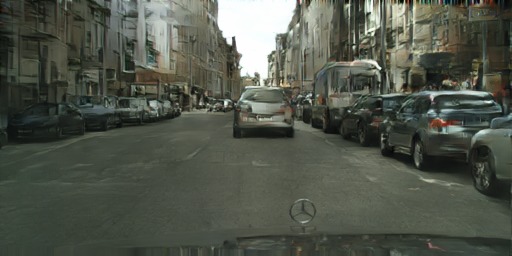} & 
\includegraphics[width=\mywidth]{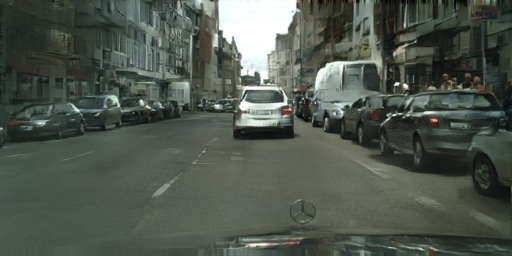} \\

\includegraphics[width=\mywidth]{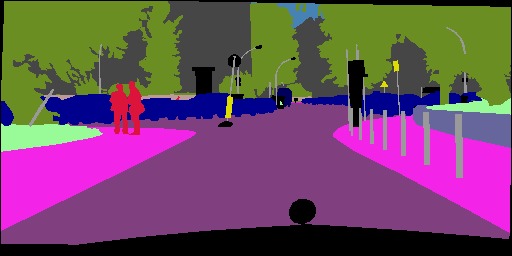} & 
\includegraphics[width=\mywidth]{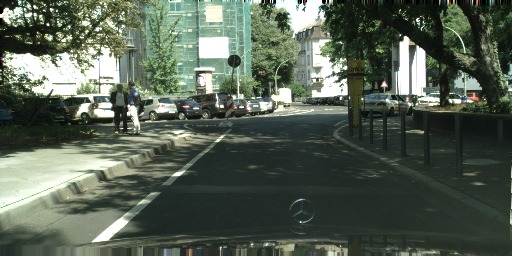} &
\includegraphics[width=\mywidth]{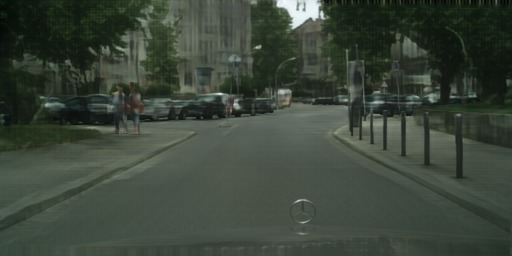} &
\includegraphics[width=\mywidth]{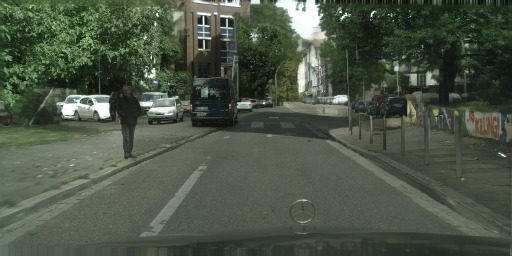} & 
\includegraphics[width=\mywidth]{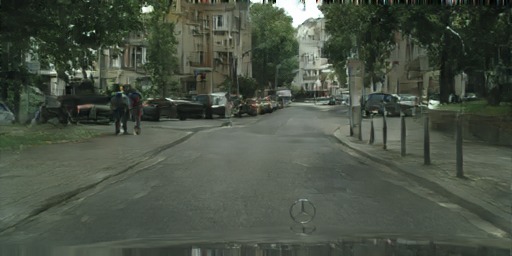} & 
\includegraphics[width=\mywidth]{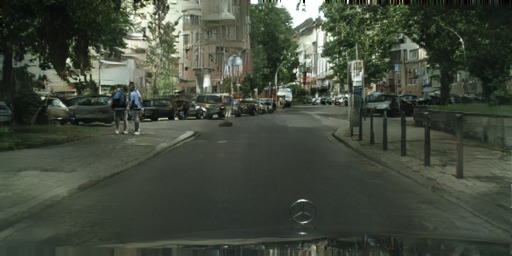}
 
\end{tabular}
    \vspace{-2mm}
	\caption{Visual comparison of semantic image synthesis results on the ADE20K outdoor and Cityscapes datasets. Our method produces realistic images while respecting the spatial semantic layout at the same time.}
	\label{fig::ade20kcomparison}
	\vspace{-2mm}
 \end{figure*}
 \egroup
 \addtolength{\tabcolsep}{4.5pt}
}

\newcommand{\lblfig}[1]{\label{fig:#1}}
\newcommand{\lblsec}[1]{\label{sec:#1}}

\newcommand{\ignorethis}[1]{}

\def\eqref#1{equation~\ref{#1}}

\def\1{\bm{1}}

\DeclareMathAlphabet{\mathsfit}{\encodingdefault}{\sfdefault}{m}{sl}
\SetMathAlphabet{\mathsfit}{bold}{\encodingdefault}{\sfdefault}{bx}{n}

\newcommand{\ignore}[1]{}

\makeatletter
\DeclareRobustCommand\onedot{\futurelet\@let@token\@onedot}
\def\@onedot{\ifx\@let@token.\else.\null\fi\xspace}
\def\eg{e.g\onedot,\xspace} 

\def\ie{i.e\onedot,\xspace}

\def\etal{\emph{et al}\onedot}
\makeatother

\cvprfinalcopy 
\def\cvprPaperID{2072} 

\ifcvprfinal\fi
\begin{document}

\title{Semantic Image Synthesis with Spatially-Adaptive Normalization}

\author{Taesung Park\textsuperscript{1,2}\thanks{} \quad Ming-Yu Liu\textsuperscript{2} \quad Ting-Chun Wang\textsuperscript{2} \quad Jun-Yan Zhu\textsuperscript{2,3} \\
\\
\textsuperscript{1}UC Berkeley \quad \textsuperscript{2}NVIDIA \quad \textsuperscript{2,3}MIT CSAIL
}

\pagestyle{plain}
\twocolumn[{
\renewcommand\twocolumn[1][]{#1}
\maketitle

\begin{center}
  \centering
  \vspace{-9mm}
  \includegraphics[width=\linewidth]{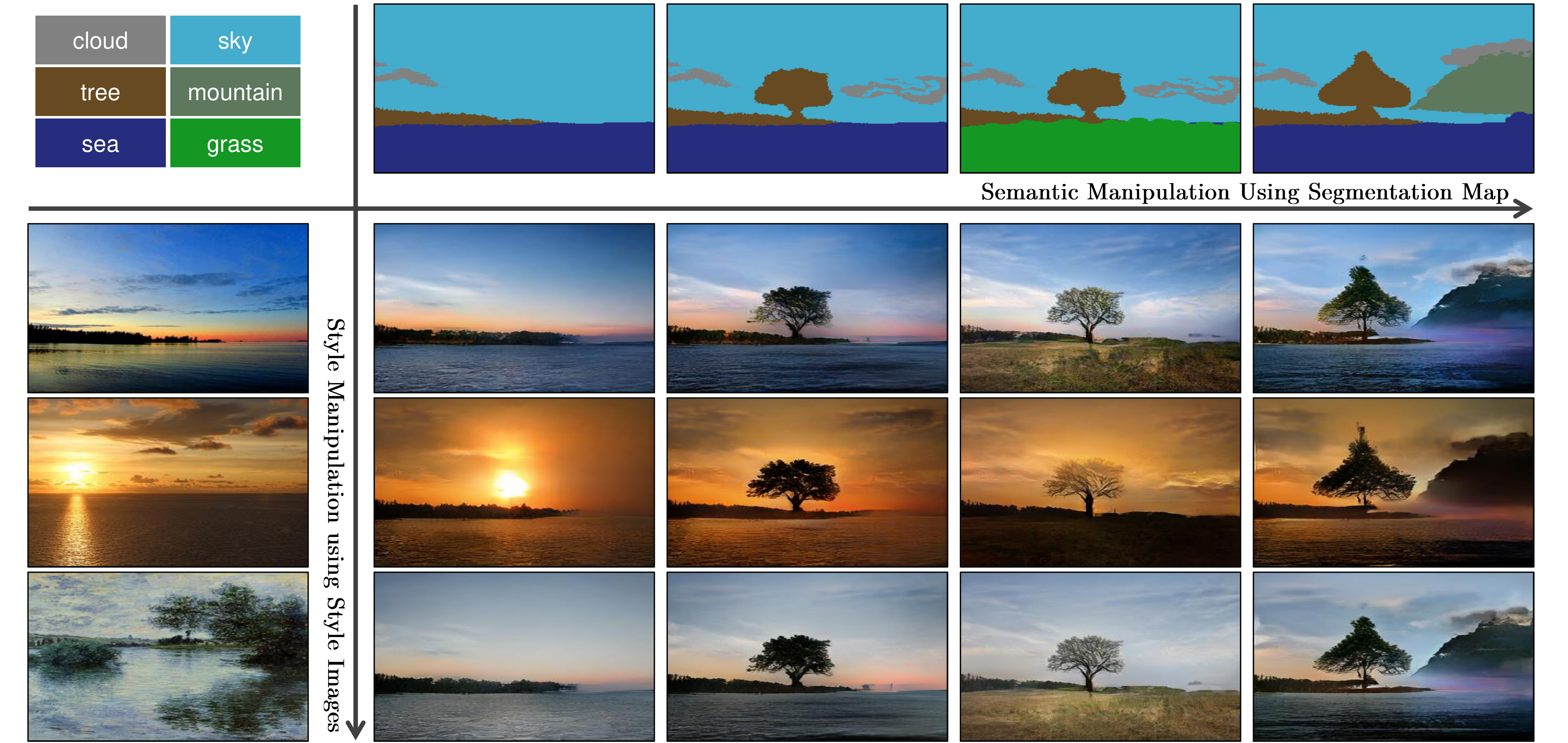}
  \vspace{-7mm}
\captionof{figure}{Our model allows user control over both semantic and style as synthesizing an image. The semantic (\eg the existence of a tree) is controlled via a label map (the top row), while the style is controlled via the reference style image (the leftmost column). Please visit our  \href{https://github.com/NVlabs/SPADE}{website} for interactive image synthesis demos.}
  \lblfig{:teaser}
\vspace{-1mm}
\end{center}}]

\maketitle
\begin{abstract}
\blfootnote{$*$Taesung Park contributed to the work during his NVIDIA internship.}
We propose spatially-adaptive normalization, a simple but effective layer for synthesizing photorealistic images given an input semantic layout. Previous methods directly feed the semantic layout as input to the deep network, which is then processed through stacks of convolution, normalization, and nonlinearity layers. We show that this is suboptimal as the normalization layers tend to ``wash away'' semantic information. To address the issue, we propose using the input layout for modulating the activations in normalization layers through a spatially-adaptive, learned transformation. Experiments on several challenging datasets demonstrate the advantage of the proposed method over existing approaches, regarding both visual fidelity and alignment with input layouts. Finally, our model allows user control over both semantic and style. Code is available at \url{https://github.com/NVlabs/SPADE}.
\end{abstract}
 \vspace{-3mm}
\section{Introduction}
 \vspace{-1mm}
Conditional image synthesis refers to the task of generating photorealistic images conditioning on certain input data. Seminal work computes the output image by stitching pieces from a single image (e.g., Image Analogies~\cite{hertzmann2001image}) or using an image collection ~\cite{johnson2006semantic,hays2007scene,lalonde2007photo,chen2009sketch2photo,Eitz:2009:PhotoSketch}. Recent methods directly learn the mapping using neural networks~\cite{isola2017image,chen2017photographic,wang2018pix2pixHD,wang2018video,zhang2017stackgan,zhang2017stackgan++,zhang2018self,brock2018large}. The latter methods are faster and require no external database of images.

We are interested in a specific form of conditional image synthesis, which is converting a semantic segmentation mask to a photorealistic image. This form has a wide range of applications such as content generation and image editing~\cite{isola2017image,chen2017photographic,wang2018pix2pixHD}. We refer to this form as semantic image synthesis. In this paper, we show that the conventional network architecture~\cite{isola2017image,wang2018pix2pixHD}, which is built by stacking convolutional, normalization, and nonlinearity layers, is at best suboptimal because their normalization layers tend to ``wash away'' information contained in the input semantic masks. To address the issue, we propose \textit{spatially-adaptive normalization}, a conditional normalization layer that modulates the activations using input semantic layouts through a spatially-adaptive, learned transformation and can effectively propagate the semantic information throughout the network.

We conduct experiments on several challenging datasets including the  COCO-Stuff~\cite{caesar2018coco,lin2014microsoft}, the ADE20K~\cite{zhou2017scene}, and the Cityscapes~\cite{cordts2016cityscapes}. We show that with the help of our spatially-adaptive normalization layer, a compact network can synthesize significantly better results compared to several state-of-the-art methods. Additionally, an extensive ablation study demonstrates the effectiveness of the proposed normalization layer against several variants for the semantic image synthesis task. Finally, our method supports multi-modal and style-guided image synthesis, enabling controllable, diverse outputs, as shown in Figure~\ref{fig::teaser}.  Also, please see our SIGGRAPH 2019 Real-Time Live \href{https://www.youtube.com/watch?v=Gz9weuemhDA&feature=youtu.be&t=2949}{demo}  and try our online  \href{https://github.com/NVlabs/SPADE}{demo} by yourself.

\section{Related Work}

\mysubsection{Deep generative models} can learn to synthesize images. Recent methods include generative adversarial networks (GANs)~\cite{goodfellow2014generative} and variational autoencoder (VAE)~\cite{kingma2013auto}. Our work is built on GANs but aims for the conditional image synthesis task. The GANs consist of a generator and a discriminator where the goal of the generator is to produce realistic images so that the discriminator cannot tell the synthesized images apart from the real ones.

\mysubsection{Conditional image synthesis} exists in many forms that differ in the type of input data. For example, class-conditional  models~\cite{odena2016conditional,miyato2018cgans,mescheder2018training,brock2018large,mirza2014conditional} learn to synthesize images given category labels. Researchers have explored various models for generating images based on text ~\cite{reed2016generative,zhang2017stackgan,xu2017attngan,hong2018inferring}. Another widely-used form is image-to-image translation based on a type of conditional GANs ~\cite{karacan2016learning,isola2017image,zhu2017unpaired,liu2017unsupervised,zhu2017toward,huang2018multimodal,Zhao2018layout,karacan2018manipulating}, where both input and output are images. Compared to earlier non-parametric methods~\cite{hertzmann2001image,johnson2006semantic,chen2009sketch2photo},  learning-based methods typically run faster during test time and produce more realistic results. In this work, we focus on converting segmentation masks to photorealistic images. We assume the training dataset contains registered segmentation masks and images. With the proposed spatially-adaptive normalization, our compact network achieves better results compared to leading methods.

\mysubsection{Unconditional normalization layers} have been an important component in modern deep networks and can be found in various classifiers, including the Local Response Normalization in the AlexNet~\cite{krizhevsky2012imagenet} and the Batch Normalization (BatchNorm) in the Inception-v2 network~\cite{ioffe2015batch}. Other popular normalization layers include the Instance Normalization (InstanceNorm)~\cite{ulyanov1607instance}, the Layer Normalization~\cite{ba2016layer}, the Group Normalization~\cite{wu2018group}, and the Weight Normalization~\cite{salimans2016weight}. We label these normalization layers as unconditional as they do not depend on external data in contrast to the conditional normalization layers discussed below.

\mysubsection{Conditional normalization layers} include the Conditional Batch Normalization (Conditional BatchNorm)~\cite{dumoulin2016learned} and Adaptive Instance Normalization (AdaIN)~\cite{huang2017adain}. Both were first used in the style transfer task and later adopted in various vision tasks~\cite{de2017modulating,perez2017learning,huang2018multimodal,miyato2018cgans,zhang2018self,mescheder2018training,wang2018recovering,chen2018on,brock2018large,karras2018style}. Different from the earlier normalization techniques, conditional normalization layers require external data and generally operate as follows. First, layer activations are normalized to zero mean and unit deviation. Then the normalized activations are \textit{denormalized} by modulating the activation using a learned affine transformation whose parameters are inferred from external data. For style transfer tasks~\cite{dumoulin2016learned,huang2017adain}, the affine parameters are used to control the global style of the output, and hence are uniform across spatial coordinates. In contrast, our proposed normalization layer applies a spatially-varying affine transformation, making it suitable for image synthesis from semantic masks. Wang~\etal proposed a closely related method for image super-resolution~\cite{wang2018recovering}. Both methods are built on spatially-adaptive modulation layers that condition on semantic inputs. While they aim to incorporate semantic information into super-resolution, our goal is to design a generator for style and semantics disentanglement. We focus on providing the semantic information in the context of modulating normalized activations. 
We use semantic maps in different scales, which enables coarse-to-fine generation. The reader is encouraged to review  their work for more details. 

\section{Semantic Image Synthesis}\label{sec::body}

Let $\mathbf{m} \in \mathbb{L}^{H\times W}$ be a semantic segmentation mask where $\mathbb{L}$ is a set of integers denoting the semantic labels, and $H$ and $W$ are the image height and width. Each entry in $\mathbf{m}$ denotes the semantic label of a pixel. We aim to learn a mapping function that can convert an input segmentation mask $\mathbf{m}$ to a photorealistic image.

\begin{figure}
 \centering
 \includegraphics[width=0.60\hsize]{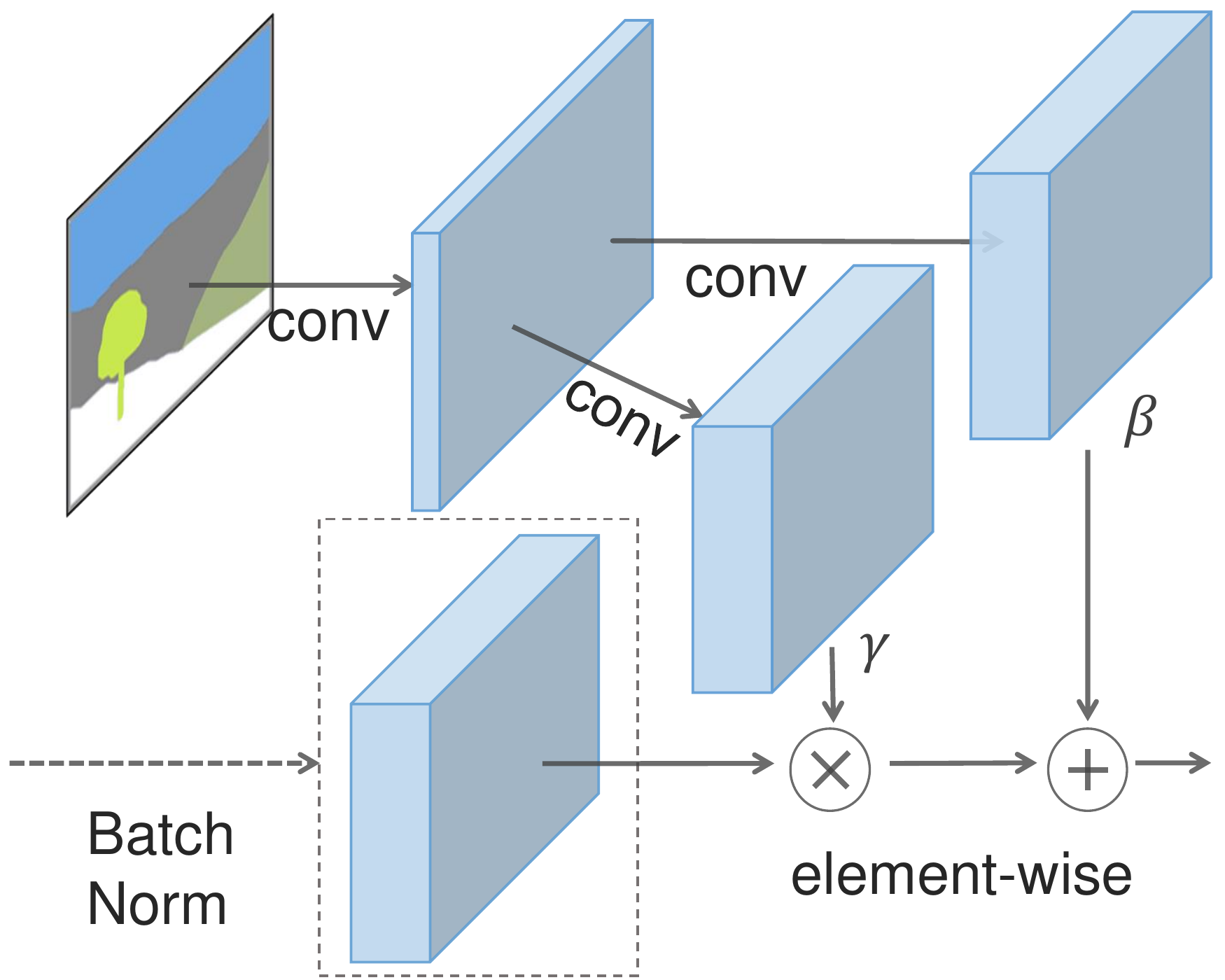}
 \vspace{-2 mm}
  \caption{In the SPADE, the mask is first projected onto an embedding space and then convolved to produce the modulation parameters $\bm{\gamma}$ and $\bm{\beta}$. Unlike prior conditional normalization methods, $\bm{\gamma}$ and $\bm{\beta}$ are not vectors, but tensors with spatial dimensions. The produced $\bm{\gamma}$ and $\bm{\beta}$ are multiplied and added to the normalized activation element-wise.}
 \label{fig::normalization}
 \vspace{-2mm}
\end{figure}

\mysubsection{Spatially-adaptive denormalization.} 
Let $\mathbf{h}^i$ denote the activations of the $i$-th layer of a deep convolutional network for a batch of $N$ samples. Let $C^i$ be the number of channels in the layer. Let $H^i$ and $W^i$ be the height and width of the activation map in the layer. We propose a new conditional normalization method called the SPatially-Adaptive (DE)normalization\footnote{Conditional normalization~\cite{dumoulin2016learned,huang2017adain} uses external data to denormalize the normalized activations; i.e., the denormalization part is conditional.} (SPADE). Similar to the Batch Normalization~\cite{ioffe2015batch}, the activation is normalized in the channel-wise manner and then modulated with learned scale and bias. Figure~\ref{fig::normalization} illustrates the SPADE design. The activation value at site $(n\in N,c \in C^i,y \in H^i,x \in W^i)$ is
\begin{equation}
\gamma^i_{c,y,x}(\mathbf{m}) \frac{h^i_{n,c,y,x}-\mu^i_c}{\sigma^i_c} + \beta^i_{c,y,x}(\mathbf{m}) \label{eqn::spade}
\end{equation}
where $h^i_{n,c,y,x}$ is the activation at the site before normalization and $\mu^i_c$ and $\sigma^i_c$ are the mean and standard deviation of the activations in channel $c$:
\begin{align}
\mu^i_c &= \frac{1}{N H^i W^i}\sum_{n,y,x} h^i_{n,c,y,x}\\
\sigma^i_c &= \sqrt{\frac{1}{N H^i W^i}\sum_{n,y,x} \Big( (h^i_{n,c,y,x})^2 - (\mu^i_c)^2 \Big) }.
\end{align}

The variables $\gamma^i_{c,y,x}(\mathbf{m})$ and $\beta^i_{c,y,x}(\mathbf{m})$ in (\ref{eqn::spade}) are the learned modulation parameters of the normalization layer. In contrast to the BatchNorm~\cite{ioffe2015batch}, they depend on the input segmentation mask and vary with respect to the location $(y, x)$. We use the symbol $\gamma^i_{c,y,x}$ and $\beta^i_{c,y,x}$ to denote the functions that convert $\mathbf{m}$ to the scaling and bias values at the site $(c,y,x)$ in the $i$-th activation map. We implement the functions $\gamma^i_{c,y,x}$ and $\beta^i_{c,y,x}$ using a simple two-layer convolutional network, whose design is in the appendix.

In fact, SPADE is related to, and is a generalization of several existing normalization layers. First, replacing the segmentation mask $\mathbf{m}$ with the image class label and making the modulation parameters spatially-invariant (\ie $\gamma^i_{c,y_1,x_1}\equiv \gamma^i_{c,y_2,x_2}$ and $\beta^i_{c,y_1,x_1}\equiv \beta^i_{c,y_2,x_2}$ for any $y_1,y_2 \in \{1,2,...,H^i\}$ and $x_1,x_2 \in \{1,2,...,W^i\}$), we arrive at the form of the Conditional BatchNorm~\cite{dumoulin2016learned}. Indeed, for any spatially-invariant conditional data, our method reduces to the Conditional BatchNorm. Similarly, we can arrive at the AdaIN~\cite{huang2017adain} by replacing $\mathbf{m}$ with a real image, making the modulation parameters spatially-invariant, and setting $N=1$. As the modulation parameters are adaptive to the input segmentation mask, the proposed SPADE is better suited for semantic image synthesis. 

\begin{figure}
	\centering
	\includegraphics[trim=0.00in 0.0in 0in 0in, width=0.3\textwidth]{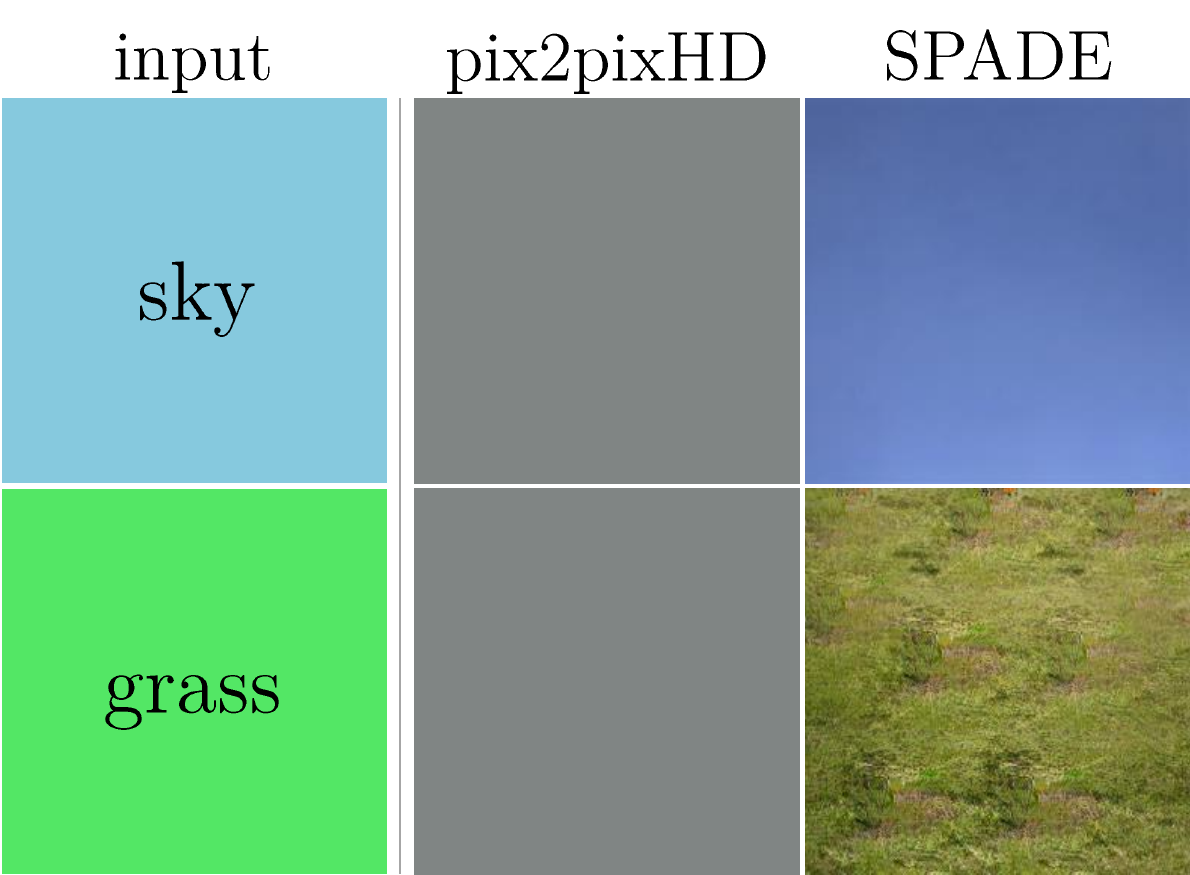}
	\vspace{-2 mm}
	\caption{Comparing results given uniform segmentation maps: while the SPADE generator produces plausible textures, the \pixtopixhd generator~\cite{wang2018pix2pixHD} produces two identical outputs due to the loss of the semantic information after the normalization layer.}
	\label{fig::uniform_label}
	\vspace{-3mm}
\end{figure}
\mysubsection{SPADE generator.} 
With the SPADE, there is no need to feed the segmentation map to the first layer of the generator, since the learned modulation parameters have encoded enough information about the label layout. Therefore, we discard encoder part of the generator, which is commonly used in recent architectures~\cite{wang2018pix2pixHD,isola2017image}. This simplification results in a more lightweight network. Furthermore, similarly to existing class-conditional generators~\cite{miyato2018cgans,zhang2018self,mescheder2018training}, the new generator can take a random vector as input, enabling a simple and natural way for multi-modal synthesis~\cite{zhu2017toward,huang2018multimodal}. 

\begin{figure*}
 \centering
 \includegraphics[width=1.0\hsize]{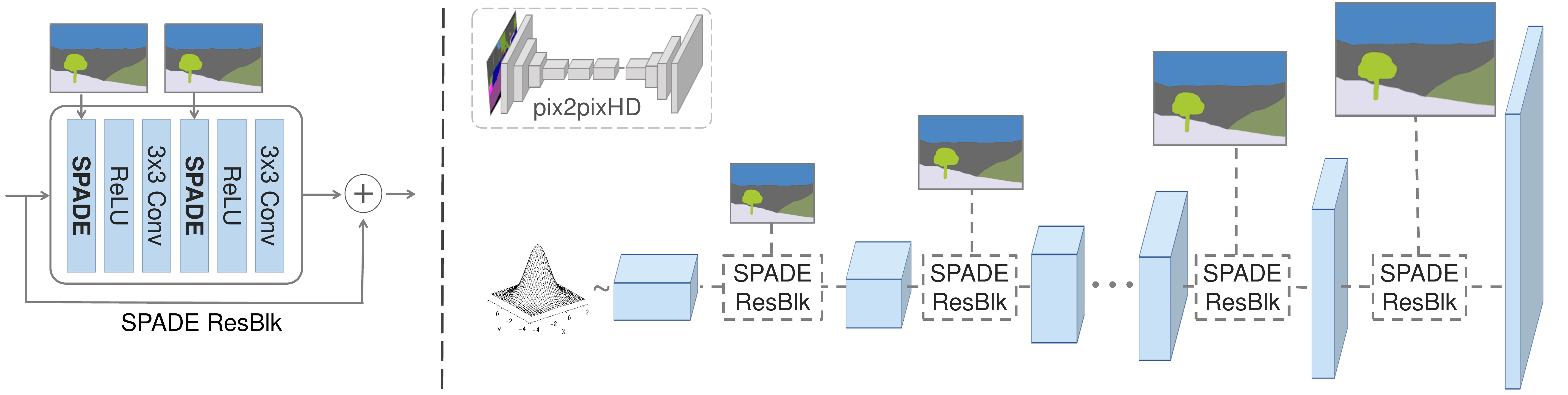}
 \vspace{-5 mm}
  \caption{In the SPADE generator, each normalization layer uses the segmentation mask to modulate the layer activations.\textit{ (left)} Structure of one residual block with the SPADE. \textit{(right) }The generator contains a series of the SPADE residual blocks with upsampling layers. Our architecture achieves better performance with a smaller number of parameters by removing the downsampling layers of leading image-to-image translation networks such as the \pixtopixhd model~\cite{wang2018pix2pixHD}. }
 \label{fig::architecture}
 \vspace{-3 mm}
\end{figure*}

Figure~\ref{fig::architecture} illustrates our generator architecture, which employs several ResNet blocks~\cite{he2016deep} with upsampling layers. The modulation parameters of all the normalization layers are learned using the SPADE. Since each residual block operates at a different scale, we downsample the semantic mask to match the spatial resolution.

We train the generator with the same multi-scale discriminator and loss function used in \pixtopixhd~\cite{wang2018pix2pixHD} except that we replace the least squared loss term~\cite{mao2017least} with the hinge loss term~\cite{lim2017geometric,miyato2018spectral,zhang2018self}. We test several ResNet-based discriminators used in recent unconditional GANs~\cite{arjovsky2017wasserstein,mescheder2018training,miyato2018cgans} but observe similar results at the cost of a higher GPU memory requirement. Adding the SPADE to the discriminator also yields a similar performance. For the loss function, we observe that removing any loss term in the \pixtopixhd loss function lead to degraded generation results.

\mysubsection{Why does the SPADE work better?} 
A short answer is that it can better preserve semantic information against common normalization layers. Specifically, while normalization layers such as the InstanceNorm~\cite{ulyanov1607instance} are essential pieces in almost all the state-of-the-art conditional image synthesis models~\cite{wang2018pix2pixHD}, they tend to wash away semantic information when applied to uniform or flat segmentation masks.

Let us consider a simple module that first applies convolution to a segmentation mask and then normalization. Furthermore, let us assume that a segmentation mask with a single label is given as input to the module (e.g., all the pixels have the same label such as sky or grass). Under this setting, the convolution outputs are again uniform, with different labels having different uniform values. Now, after we apply InstanceNorm to the output, the normalized activation will become all zeros no matter what the input semantic label is given. Therefore, semantic information is totally lost. This limitation applies to a wide range of generator architectures, including \pixtopixhd and its variant that concatenates the semantic mask at all intermediate layers, as long as a network applies convolution and then normalization to the semantic mask. In Figure~\ref{fig::uniform_label}, we empirically show this is precisely the case for \pixtopixhd. Because a segmentation mask consists of a few uniform regions in general,  the issue of information loss emerges when applying normalization.

In contrast, the segmentation mask in the SPADE Generator is fed through spatially adaptive modulation \textit{without} normalization. Only activations from the previous layer are normalized. Hence, the SPADE generator can better preserve semantic information. It enjoys the benefit of normalization without losing the semantic input information.

\addtolength{\tabcolsep}{-4.5pt}    
\setlength{\mywidth}{0.1932\textwidth}
\bgroup
\def\arraystretch{0.5}
\begin{figure*}[]
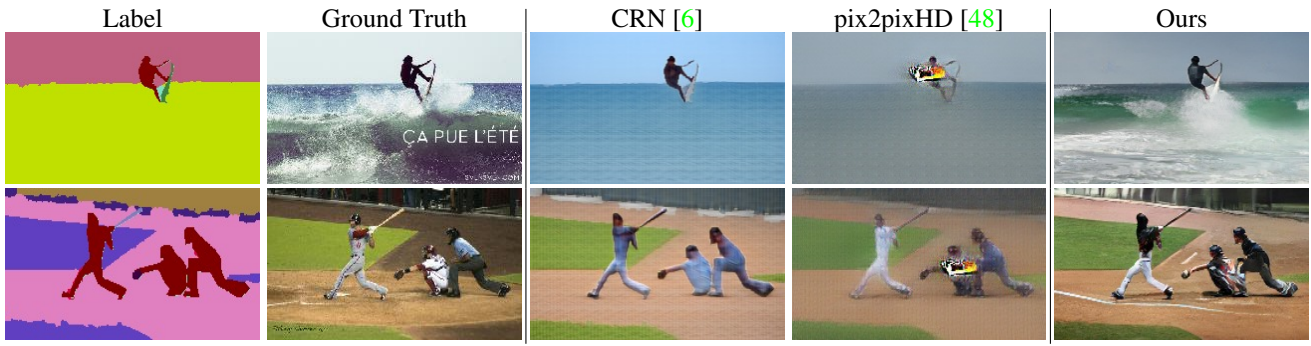

\begin{tabular} {cc|cc|c}
 Label & Ground Truth & CRN~\cite{chen2017photographic} &  \pixtopixhd~\cite{wang2018pix2pixHD} & Ours \\
 \includegraphics[width=\mywidth, height=0.79in]{images/coco/label/000000263403.jpg} & \includegraphics[width=\mywidth, height=0.79in]{images/coco/gt/000000263403.jpg} &
 \includegraphics[width=\mywidth, height=0.79in]{images/coco/crn/000000263403.jpg} &   \includegraphics[width=\mywidth, height=0.79in]{images/coco/pix2pixhd/000000263403.jpg} &  \includegraphics[width=\mywidth, height=0.79in]{images/coco/ours/000000263403.jpg} \\
 \includegraphics[width=\mywidth, height=0.79in]{images/coco/label/000000129945.jpg} & \includegraphics[width=\mywidth, height=0.79in]{images/coco/gt/000000129945.jpg} &
 \includegraphics[width=\mywidth, height=0.79in]{images/coco/crn/000000129945.jpg} &   \includegraphics[width=\mywidth, height=0.79in]{images/coco/pix2pixhd/000000129945.jpg} &  \includegraphics[width=\mywidth, height=0.79in]{images/coco/ours/000000129945.jpg} \\
\end{tabular}
\vspace{-2mm}
	\caption{Visual comparison of semantic image synthesis results on the COCO-Stuff dataset. Our method successfully synthesizes realistic details from semantic labels. }
	\label{fig::cococomparison}	
\vspace{-4mm}	
 \end{figure*}
 \egroup
 \addtolength{\tabcolsep}{4.5pt}

\addtolength{\tabcolsep}{-4.5pt}    

\setlength{\mywidth}{0.16\textwidth}

\bgroup
\def\arraystretch{0.5}
\begin{figure*}[]
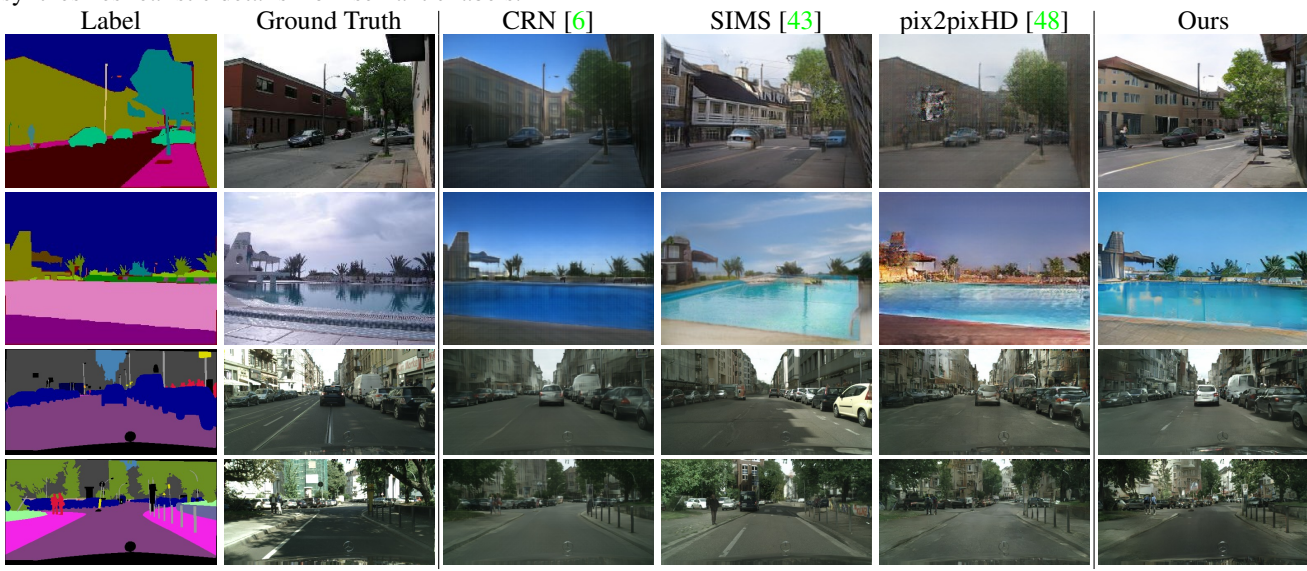

\begin{tabular} {cc|ccc|c}
 Label & Ground Truth & CRN~\cite{chen2017photographic} & SIMS~\cite{qi2018semi} & \pixtopixhd~\cite{wang2018pix2pixHD} & Ours \\
 
\includegraphics[width=\mywidth, height=0.80in]{images/ade20k/label/ADE_val_00000786.jpg} & \includegraphics[width=\mywidth, height=0.80in]{images/ade20k/gt/ADE_val_00000786.jpg} &
\includegraphics[width=\mywidth, height=0.80in]{images/ade20k/crn/ADE_val_00000786.jpg} &   \includegraphics[width=\mywidth, height=0.80in]{images/ade20k/sims/ADE_val_00000786.jpg} &
\includegraphics[width=\mywidth, height=0.80in]{images/ade20k/pix2pixhd/ADE_val_00000786.jpg} &  \includegraphics[width=\mywidth, height=0.80in]{images/ade20k/ours/ADE_val_00000786.jpg} \\

\includegraphics[width=\mywidth, height=0.80in]{images/ade20k/label/ADE_val_00001880.jpg} & \includegraphics[width=\mywidth, height=0.80in]{images/ade20k/gt/ADE_val_00001880.jpg} &
\includegraphics[width=\mywidth, height=0.80in]{images/ade20k/crn/ADE_val_00001880.jpg} &  \includegraphics[width=\mywidth, height=0.80in]{images/ade20k/sims/ADE_val_00001880.jpg} &
\includegraphics[width=\mywidth, height=0.80in]{images/ade20k/pix2pixhd/ADE_val_00001880.jpg} &  \includegraphics[width=\mywidth, height=0.80in]{images/ade20k/ours/ADE_val_00001880.jpg} \\

\includegraphics[width=\mywidth]{images/cityscapes/label/frankfurt_000001_032711_gtFine_color.jpg} & 
\includegraphics[width=\mywidth]{images/cityscapes/gt/frankfurt_000001_032711_leftImg8bit.jpg} &
\includegraphics[width=\mywidth]{images/cityscapes/crn/frankfurt_000001_032711_leftImg8bit.jpg} &
\includegraphics[width=\mywidth]{images/cityscapes/sims/frankfurt_000001_032711_leftImg8bit.jpg} & 
\includegraphics[width=\mywidth]{images/cityscapes/pix2pixhd/frankfurt_000001_032711_leftImg8bit.jpg} & 
\includegraphics[width=\mywidth]{images/cityscapes/ours/frankfurt_000001_032711_leftImg8bit.jpg} \\

\includegraphics[width=\mywidth]{images/cityscapes/label/frankfurt_000001_046272_gtFine_color.jpg} & 
\includegraphics[width=\mywidth]{images/cityscapes/gt/frankfurt_000001_046272_leftImg8bit.jpg} &
\includegraphics[width=\mywidth]{images/cityscapes/crn/frankfurt_000001_046272_leftImg8bit.jpg} &
\includegraphics[width=\mywidth]{images/cityscapes/sims/frankfurt_000001_046272_leftImg8bit.jpg} & 
\includegraphics[width=\mywidth]{images/cityscapes/pix2pixhd/frankfurt_000001_046272_leftImg8bit.jpg} & 
\includegraphics[width=\mywidth]{images/cityscapes/ours/frankfurt_000001_046272_leftImg8bit.jpg}
 
\end{tabular}
    \vspace{-2mm}
	\caption{Visual comparison of semantic image synthesis results on the ADE20K outdoor and Cityscapes datasets. Our method produces realistic images while respecting the spatial semantic layout at the same time.}
	\label{fig::ade20kcomparison}
	\vspace{-2mm}
 \end{figure*}
 \egroup
 \addtolength{\tabcolsep}{4.5pt}

\mysubsection{Multi-modal synthesis.} 
By using a random vector as the input of the generator, our architecture provides a simple way for multi-modal synthesis~\cite{zhu2017toward,huang2018multimodal}. Namely, one can attach an encoder that processes a real image into a random vector, which will be then fed to the generator. The encoder and generator form a VAE~\cite{kingma2013auto}, in which the encoder tries to capture the style of the image, while the generator combines the encoded style and the segmentation mask information via the SPADEs to reconstruct the original image. The encoder also serves as a style guidance network at test time to capture the style of target images, as used in Figure~\ref{fig::teaser}. For training, we add a KL-Divergence loss term~\cite{kingma2013auto}.

\section{Experiments}

\begin{table*}[tb!]
	\vspace{-1mm}
	\centering
	\small
    \begin{tabular}{r||c|c|c|c|c|c|c|c|c|c|c|c}
& \multicolumn{3}{c}{COCO-Stuff}  & \multicolumn{3}{|c}{ADE20K} & \multicolumn{3}{|c}{ADE20K-outdoor} & \multicolumn{3}{|c}{Cityscapes} \\
\cline{2-13}
Method & mIoU & accu & FID & mIoU & accu & FID & mIoU & accu & FID & mIoU & accu & FID \\
\Xhline{2\arrayrulewidth}
CRN~\cite{chen2017photographic} & 23.7& 40.4& 70.4& 22.4& 68.8& 73.3& 16.5& 68.6& 99.0& 52.4& 77.1& 104.7 \\
SIMS~\cite{qi2018semi} &  N/A& N/A& N/A& N/A& N/A& N/A& 13.1& 74.7& 67.7& 47.2& 75.5& {\bf49.7}\\
\pixtopixhd~\cite{wang2018pix2pixHD} & 14.6& 45.8& 111.5& 20.3& 69.2& 81.8& 17.4& 71.6& 97.8& 58.3& 81.4& 95.0 \\
{\bf \Ours } &  {\bf37.4} &  {\bf 67.9} &  {\bf 22.6} &  {\bf 38.5} &  {\bf 79.9} &  {\bf 33.9} &  {\bf 30.8} &  {\bf 82.9} &  {\bf 63.3} &  {\bf 62.3} &  {\bf 81.9} & 71.8 \\
	\end{tabular}\vspace{-2mm}
	\caption{Our method outperforms the current leading methods in semantic segmentation (mIoU and accu) and FID~\cite{heusel2017gans} scores on all the benchmark datasets. For the mIoU and accu, higher is better. For the FID, lower is better.}
	\label{tbl::comparison}		
 \vspace{-3mm}
 \end{table*}
\begin{figure*}[]
  \centering
  \includegraphics[width=0.95\linewidth]{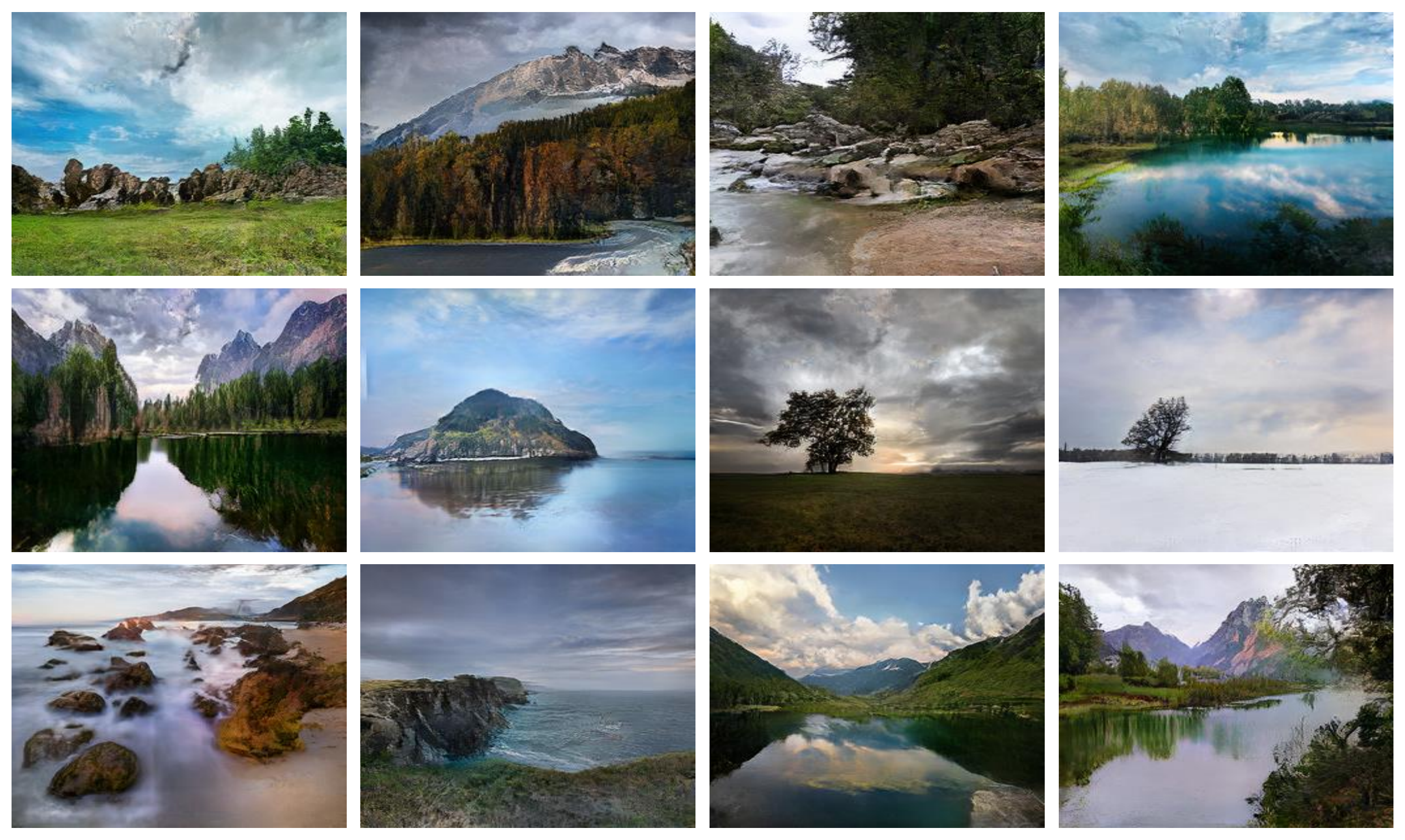}
  \vspace{-4mm}
  \caption{Semantic image synthesis results on the Flickr Landscapes dataset. The images were generated from semantic layout of photographs on the Flickr website.}  \label{fig::flickr}
  \vspace{-4mm}
\end{figure*}

\mysubsection{Implementation details.} We apply the Spectral Norm~\cite{miyato2018spectral} to all the layers in both generator and discriminator. The learning rates for the generator and discriminator are $0.0001$ and $0.0004$, respectively~\cite{heusel2017gans}. We use the ADAM solver~\cite{kingma2015adam} with $\beta_1=0$ and $\beta_2=0.999$. All the experiments are conducted on an NVIDIA DGX1 with 8 32GB V100 GPUs. We use synchronized BatchNorm, \ie these statistics are collected from all the GPUs.

\mysubsection{Datasets.} We conduct experiments on several datasets.
\begin{itemize}[leftmargin=*,noitemsep,topsep=0pt]
\item \textit{COCO-Stuff}~\cite{caesar2018coco} is derived from the COCO dataset~\cite{lin2014microsoft}. It has $118,000$ training images and $5,000$ validation images captured from diverse scenes. It has $182$ semantic classes. Due to its vast diversity, existing image synthesis models perform poorly on this dataset.
\item \textit{ADE20K}~\cite{zhou2017scene} consists of $20,210$ training and $2,000$ validation images. Similarly to the COCO, the dataset contains challenging scenes with 150 semantic classes. 
\item \textit{ADE20K-outdoor} is a subset of the ADE20K dataset that only contains outdoor scenes, used in Qi \etal ~\cite{qi2018semi}. 
\item \textit{Cityscapes} dataset~\cite{cordts2016cityscapes} contains street scene images in German cities. The training and validation set sizes are $3,000$ and $500$, respectively. Recent work has achieved photorealistic semantic image synthesis results~\cite{wang2018video,qi2018semi} on the Cityscapes dataset. 
\item \textit{Flickr Landscapes.} We collect $41,000$ photos from Flickr and use $1,000$ samples for the validation set. To avoid expensive manual annotation, we use a well-trained DeepLabV2~\cite{chen2018deeplab} to compute input segmentation masks.
\end{itemize}
We train the competing semantic image synthesis methods on the same training set and report their results on the same validation set for each dataset.

\mysubsection{Performance metrics.} We adopt the evaluation protocol from previous work~\cite{chen2017photographic,wang2018pix2pixHD}. Specifically, we run a semantic segmentation model on the synthesized images and compare how well the predicted segmentation mask matches the ground truth input. 
Intuitively, if the output images are realistic, a well-trained semantic segmentation model should be able to predict the ground truth label. For measuring the segmentation accuracy, we use both the mean Intersection-over-Union (mIoU) and the pixel accuracy (accu). We use the state-of-the-art segmentation networks for each dataset: DeepLabV2~\cite{chen2018deeplab,kazuto2018} for COCO-Stuff, UperNet101~\cite{xiao2018unified} for ADE20K, and DRN-D-105~\cite{yu2017dilated} for Cityscapes. In addition to the mIoU and the accu segmentation performance metrics, we use the Fr\'echet Inception Distance (FID)~\cite{heusel2017gans} to measure the distance between the distribution of synthesized results and the distribution of real images.

\mysubsection{Baselines.} We compare our method with 3 leading semantic image synthesis models: the \pixtopixhd model~\cite{wang2018pix2pixHD}, the cascaded refinement network (\crn)~\cite{chen2017photographic}, and the semi-parametric image synthesis method (\sims)~\cite{qi2018semi}. The \pixtopixhd is the current state-of-the-art GAN-based conditional image synthesis framework. The \crn uses a deep network that repeatedly refines the output from low to high resolution, while the \sims takes a semi-parametric approach that composites real segments from a training set and refines the boundaries. Both the \crn and \sims are mainly trained using image reconstruction loss. For a fair comparison, we train the \crn and \pixtopixhd models using the implementations provided by the authors. As image synthesis using the \sims requires many queries to the training dataset, it is computationally prohibitive for a large dataset such as the COCO-stuff and the full ADE20K. Therefore, we use the results provided by the authors when available.

\mysubsection{Quantitative comparisons.} As shown in Table~\ref{tbl::comparison}, our method outperforms the current state-of-the-art methods by a large margin in all the datasets. For the COCO-Stuff, our method achieves an mIoU score of $35.2$, which is about $1.5$ times better than the previous leading method. Our FID is also $2.2$ times better than the previous leading method. We note that the \sims model produces a lower FID score but has poor segmentation performances on the Cityscapes dataset. This is because the \sims synthesizes an image by first stitching image patches from the training dataset. As using the real image patches, the resulting image distribution can better match the distribution of real images. However, because there is no guarantee that a perfect query (\eg a person in a particular pose) exists in the dataset, it tends to copy objects that do not match the input segments.

\mysubsection{Qualitative results.} In Figures~\ref{fig::cococomparison}~and~\ref{fig::ade20kcomparison}, we provide qualitative comparisons of the competing methods. We find that our method produces results with much better visual quality and fewer visible artifacts, especially for diverse scenes in the COCO-Stuff and ADE20K dataset. When the training dataset size is small, the \sims model also renders images with good visual quality. However, the depicted content often deviates from the input segmentation mask (\eg the shape of the swimming pool in the second row of Figure~\ref{fig::ade20kcomparison}). 

In Figures~\ref{fig::flickr} and~\ref{fig::coco}, we show more example results from the Flickr Landscape and COCO-Stuff datasets. The proposed method can generate diverse scenes with high image fidelity. More results are included in the appendix.

\addtolength{\tabcolsep}{-5pt}    

\newlength{\mywidthcoco}
\setlength{\mywidthcoco}{0.193\textwidth}
\newlength{\myheightcoco}

\bgroup
\def\arraystretch{0.5}
\begin{figure*}[]
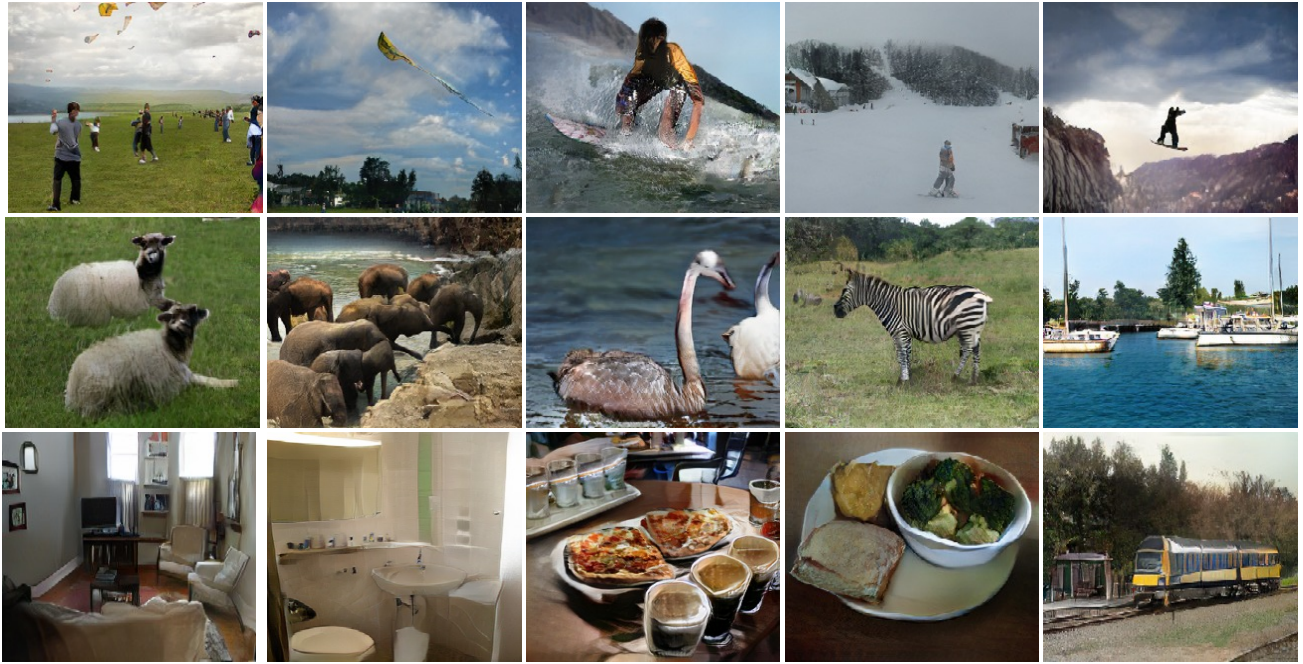

\begin{tabular}{ccccc}
\setlength{\myheightcoco}{1.1in}
\setlength{\myheightcoco}{1.1in}
 \includegraphics[width=\mywidthcoco, height=1.1in]{images/coco/ours/000000193245.jpg} & \includegraphics[width=\mywidthcoco, height=1.1in]{images/coco/ours/000000024027.jpg} & \includegraphics[width=\mywidthcoco, height=1.1in]{images/coco/ours/000000051314.jpg} & \includegraphics[width=\mywidthcoco, height=1.1in]{images/coco/ours/000000445602.jpg} & \includegraphics[width=\mywidthcoco, height=1.1in]{images/coco/ours/000000407650.jpg} \\
  \setlength{\myheightcoco}{1.1in}
 \includegraphics[width=\mywidthcoco,height=1.1in]{images/coco/ours/000000547383.jpg} &
 \includegraphics[width=\mywidthcoco,height=1.1in]{images/coco/ours/000000383921.jpg} &
 \includegraphics[width=\mywidthcoco,height=1.1in]{images/coco/ours/000000436551.jpg} &
 \includegraphics[width=\mywidthcoco,height=1.1in]{images/coco/ours/000000250758.jpg} &
 \includegraphics[width=\mywidthcoco,height=1.1in]{images/coco/ours/000000490470.jpg} \\
 \includegraphics[width=\mywidthcoco,height=1.2in]{images/coco/ours/000000139684.jpg} &
 \includegraphics[width=\mywidthcoco,height=1.2in]{images/coco/ours/000000062025.jpg} &
 \includegraphics[width=\mywidthcoco,height=1.2in]{images/coco/ours/000000074733.jpg} &
 \includegraphics[width=\mywidthcoco,height=1.2in]{images/coco/ours/000000030785.jpg} &
 \includegraphics[width=\mywidthcoco,height=1.2in]{images/coco/ours/000000206994.jpg} \\
\end{tabular}
\vspace{-3mm}
	\caption{Semantic image synthesis results on COCO-Stuff. Our method successfully generates realistic images in diverse scenes ranging from animals to sports activities.} \label{fig::coco}		
  	\vspace{-4mm}
 \end{figure*}
 \egroup
 \addtolength{\tabcolsep}{5pt}

\begin{table}[t!]
	\centering
	\small
    \begin{tabular}{r||c|c|c}
		\multirow{2}{*}{Dataset} & Ours vs. & Ours vs. & Ours vs.\\
		& \crn & \pixtopixhd & \sims\\		
		\Xhline{2\arrayrulewidth}
		COCO-Stuff & 79.76 & 86.64 & N/A\\
		ADE20K & 76.66 & 83.74 & N/A\\
		ADE20K-outdoor & 66.04	& 79.34 &	85.70 \\
	Cityscapes & 63.60 & 53.64 & 51.52 \\
	\end{tabular}\vspace{-3mm}
	\caption{User preference study. The numbers indicate the percentage of users who favor the results of the proposed method over those of the competing method.}\label{tbl::user_study}		
 \vspace{-3mm}
 \end{table}

\mysubsection{Human evaluation.} We use the Amazon Mechanical Turk (AMT) to compare the perceived visual fidelity of our method against existing approaches. Specifically, we give the AMT workers an input segmentation mask and two synthesis outputs from different methods and ask them to choose the output image that looks more like a corresponding image of the segmentation mask. The workers are given unlimited time to make the selection. For each comparison, we randomly generate $500$ questions for each dataset, and each question is answered by $5$ different workers. For quality control, only workers with a lifetime task approval rate greater than $98\%$ can participate in our study. 

Table~\ref{tbl::user_study} shows the evaluation results. We find that users strongly favor our results on all the datasets, especially on the challenging COCO-Stuff and ADE20K datasets. For the Cityscapes, even when all the competing methods achieve high image fidelity, users still prefer our results.

\bgroup
\def\arraystretch{1.0}
\addtolength{\tabcolsep}{-2.5pt}  
\begin{table}[!t]
	\centering
	\small
    \begin{tabular}{r||c||c|c|c}
		Method & \#param & COCO. & ADE. & City.  \\
		\Xhline{2\arrayrulewidth}
		{\bf decoder w/ SPADE (\Ours)}  &	96M	& \textbf{35.2}	& 38.5	& 62.3 \\
		{\bf compact decoder w/ SPADE }  &	61M	& \textbf{35.2} & 38.0 & \textbf{62.5} \\
        decoder w/ Concat  &	79M	& 31.9	& 33.6	& 61.1 \\
        \hline
        {\bf \pixtopixplusplus w/ SPADE}	& 237M	&34.4	&\textbf{39.0}	& 62.2 \\
        \pixtopixplusplus w/ Concat	& 195M	& 32.9	& 38.9	& 57.1 \\
        \pixtopixplusplus 	& 183M	& 32.7 &	38.3	&58.8 \\
        compact \pixtopixplusplus & 103M &	31.6 & 37.3	& 57.6 \\
        \pixtopixhd ~\cite{wang2018pix2pixHD}	& 183M	& 14.6	& 20.3	& 58.3 \\
	\end{tabular}\vspace{-2mm}
	\caption{The mIoU scores are boosted when the SPADE is used, for both the decoder architecture  (Figure \ref{fig::architecture}) and encoder-decoder architecture of \pixtopixplusplus (our improved baseline over \pixtopixhd~\cite{wang2018pix2pixHD}). On the other hand, simply concatenating semantic input at every layer fails to do so. Moreover, our compact model with smaller depth at all layers outperforms all the baselines.}
	\label{tbl::effectiveness}		
 \vspace{-3mm}
 \end{table}
 \addtolength{\tabcolsep}{2.5pt}  
 \egroup
 
\begin{table}[!tbh]
	\centering
	\small
    \begin{tabular}{r||c|c|c}
Method  & COCO & ADE20K & Cityscapes  \\
\Xhline{2\arrayrulewidth}
\textbf{segmap input} & 35.2 & 38.5 & 62.3 \\
\textbf{random input} & 35.3 & 38.3 & 61.6 \\
\hline
kernelsize 5x5 & 35.0 & 39.3 & 61.8 \\
\textbf{kernelsize 3x3} & 35.2 & 38.5 & 62.3\\
kernelsize 1x1 & 32.7 & 35.9 & 59.9\\
\hline
\#params 141M  & 35.3 & 38.3 & 62.5\\
\textbf{\#params 96M} & 35.2 & 38.5 & 62.3\\
\#params 61M & 35.2 & 38.0 & 62.5\\
\hline
\textbf{Sync BatchNorm} & 35.0 & 39.3 & 61.8\\
BatchNorm & 33.7 & 37.9 & 61.8\\
InstanceNorm & 33.9 & 37.4 & 58.7\\

\end{tabular}\vspace{-2mm}
\caption{The SPADE generator works with different configurations. We change the input of the generator, the convolutional kernel size acting on the segmentation map, the capacity of the network, and the parameter-free normalization method. The settings used in the paper are boldfaced.}\label{tbl::variation}		
 \vspace{-3mm}
 \end{table}
 
\begin{figure*}[]
  \centering
  \includegraphics[width=\linewidth]{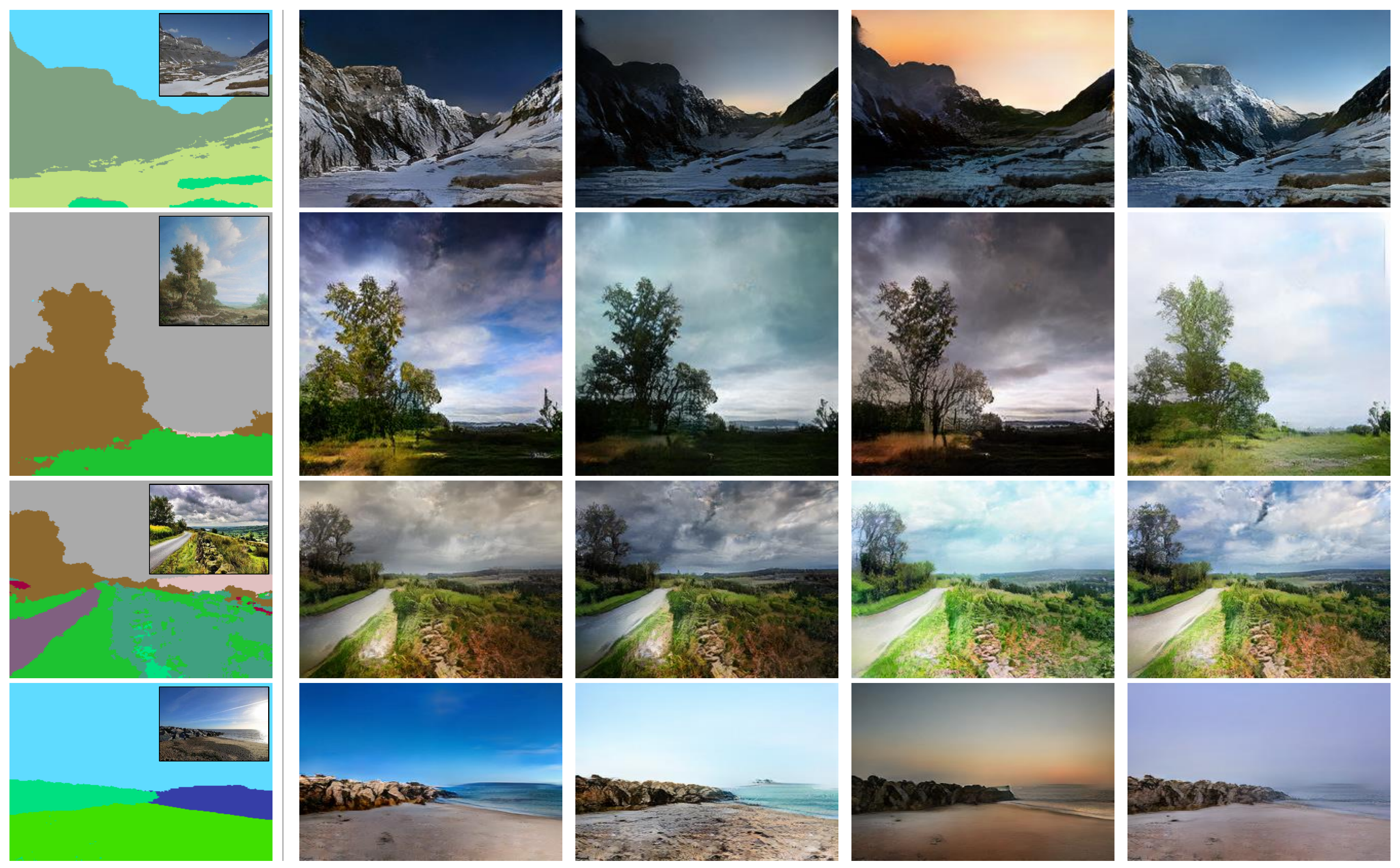}
  \vspace{-7mm}
\captionof{figure}{Our model attains multimodal synthesis capability when trained with the image encoder. During deployment, by using different random noise, our model synthesizes outputs with diverse appearances but all having the same semantic layouts depicted in the input mask. For reference, the ground truth image is shown inside the input segmentation mask.} \label{fig::multimodal}
 \vspace{-3mm}
\end{figure*}

\mysubsection{Effectiveness of the SPADE.} For quantifying importance of the SPADE, we introduce a strong baseline called \pixtopixplusplus, which combines all the techniques we find useful for enhancing the performance of \pixtopixhd except the SPADE. We also train models that receive the segmentation mask input at all the intermediate layers via feature concatenation in the channel direction, which is termed as \pixtopixplusplus w/ Concat. Finally, the model that combines the strong baseline with the SPADE is denoted as \pixtopixplusplus w/ SPADE. 

As shown in Table \ref{tbl::effectiveness}, the architectures with the proposed SPADE consistently outperforms its counterparts, in both the decoder-style architecture described in Figure \ref{fig::architecture} and more traditional encoder-decoder architecture used in the \pixtopixhd. We also find that concatenating segmentation masks at all intermediate layers, a reasonable alternative to the SPADE, does not achieve the same performance as SPADE. Furthermore, the decoder-style SPADE generator works better than the strong baselines even with a smaller number of parameters. 

\mysubsection{\bf Variations of SPADE generator.} Table \ref{tbl::variation} reports the performance of several variations of our generator. First, we compare two types of input to the generator where one is the random noise while the other is the downsampled segmentation map. We find that both of the variants render similar performance and conclude that the modulation by SPADE alone provides sufficient signal about the input mask. Second, we vary the type of parameter-free normalization layers before applying the modulation parameters. We observe that the SPADE works reliably across different normalization methods. Next, we vary the convolutional kernel size acting on the label map, and find that kernel size of 1x1 hurts performance, likely because it prohibits utilizing the context of the label. Lastly, we modify the capacity of the generator by changing the number of convolutional filters. We present more variations and ablations in the appendix.

\mysubsection{Multi-modal synthesis.} In Figure~\ref{fig::multimodal}, we show the multimodal image synthesis results on the Flickr Landscape dataset. For the same input segmentation mask, we sample different noise inputs to achieve different outputs. More results are included in the appendix.

\mysubsection{Semantic manipulation and guided image synthesis.} In Figure~\ref{fig::teaser}, we show an application where a user draws different segmentation masks, and our model renders the corresponding landscape images. Moreover, our model allows users to choose an external style image to control the global appearances of the output image. We achieve it by replacing the input noise with the embedding vector of the style image computed by the image encoder.

\section{Conclusion}\lblsec{conclusion}

We have proposed the spatially-adaptive normalization, which utilizes the input semantic layout while performing the affine transformation in the normalization layers. The proposed normalization leads to the first semantic image synthesis model that can produce photorealistic outputs for diverse scenes including indoor, outdoor, landscape, and street scenes. We further demonstrate its application for multi-modal synthesis and guided image synthesis. 

\mysubsection{Acknowledgments}. We thank Alexei A. Efros, Bryan Catanzaro, Andrew Tao, and Jan Kautz for insightful advice. We thank Chris Hebert, Gavriil Klimov, and Brad Nemire for their help in constructing the demo apps. Taesung Park contributed to the work during his internship at NVIDIA. His Ph.D. is supported by a Samsung Scholarship.

\clearpage
\balance
{\small
\bibliographystyle{ieee}
\bibliography{egbib}
}

\clearpage
\appendix
\section{Additional Implementation Details}

\begin{figure}[!b]
\centering
\includegraphics[trim=0.00in 0.0in 0in 0in, width=0.3\textwidth]{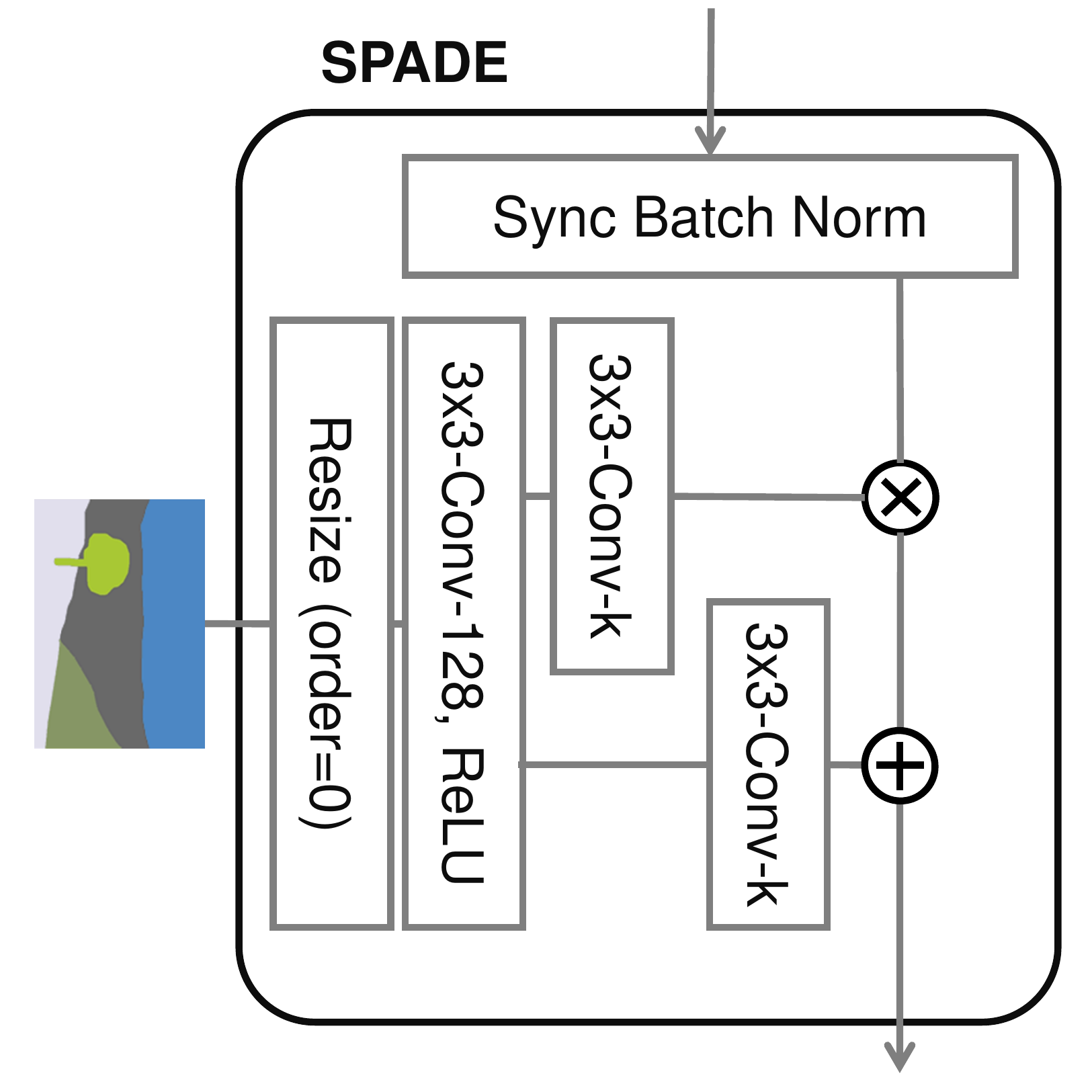}
  \vspace{-2mm}
  \caption{SPADE Design. The term 3x3-Conv-k denotes a 3-by-3 convolutional layer with $k$ convolutional filters. The segmentation map is resized to match the resolution of the corresponding feature map using nearest-neighbor downsampling.}
  \label{fig::supp_spade_arch}
  \vspace{5mm}
\includegraphics[trim=0.00in 0.0in 0in 0in, width=0.3\textwidth]{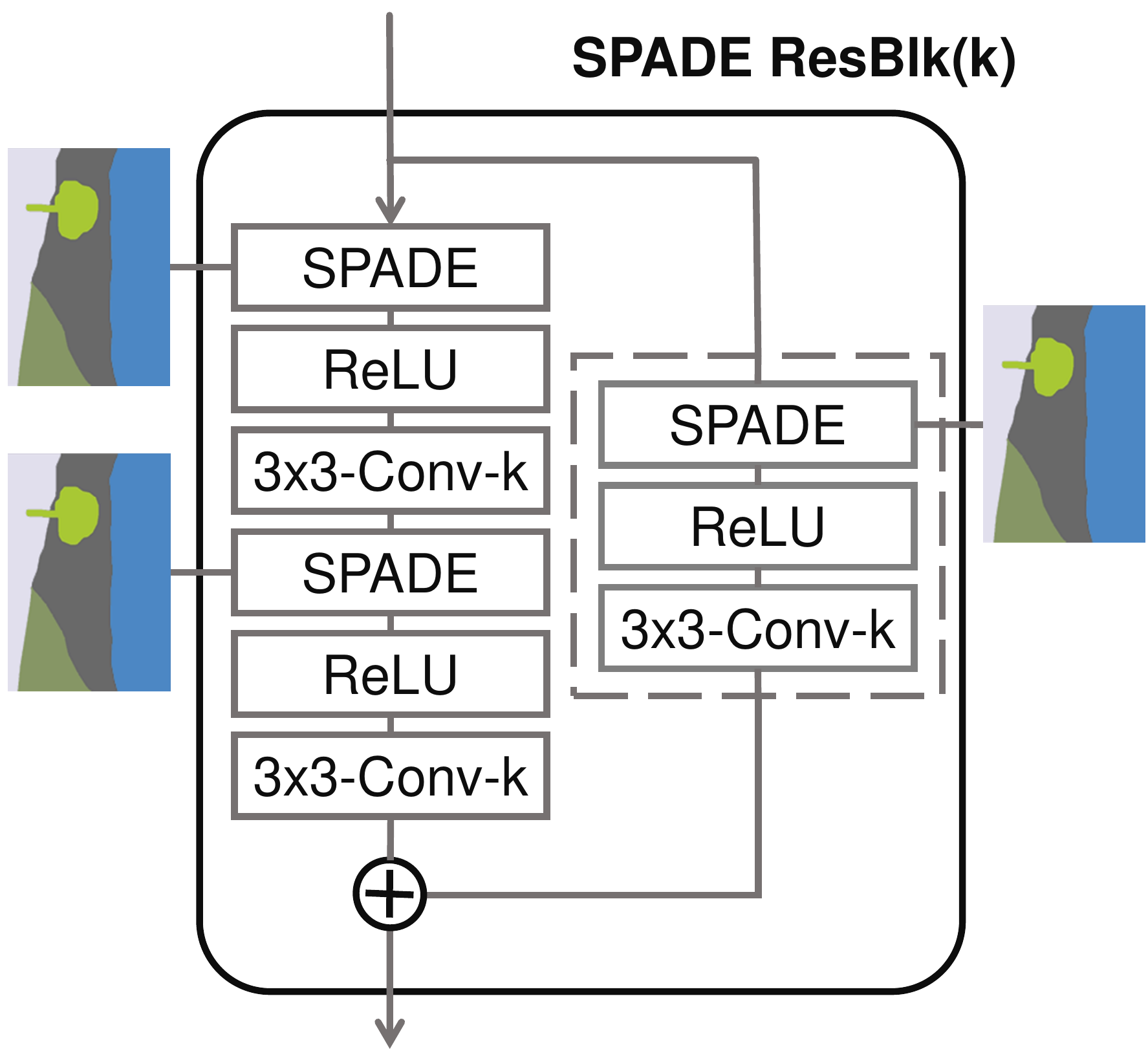}
  \vspace{-2mm}
  \caption{SPADE ResBlk. The residual block design largely follows that in Mescheder~\etal~\cite{mescheder2018training} and Miyato~\etal ~\cite{miyato2018cgans}. We note that for the case that the number of channels before and after the residual block is different, the skip connection is also learned (dashed box in the figure).}
  \label{fig::supp_spade_resblk_arch}
\end{figure}

\begin{figure}[!tbh]
\centering
\includegraphics[trim=0.00in 0.0in 0in 0in, width=0.35\textwidth]{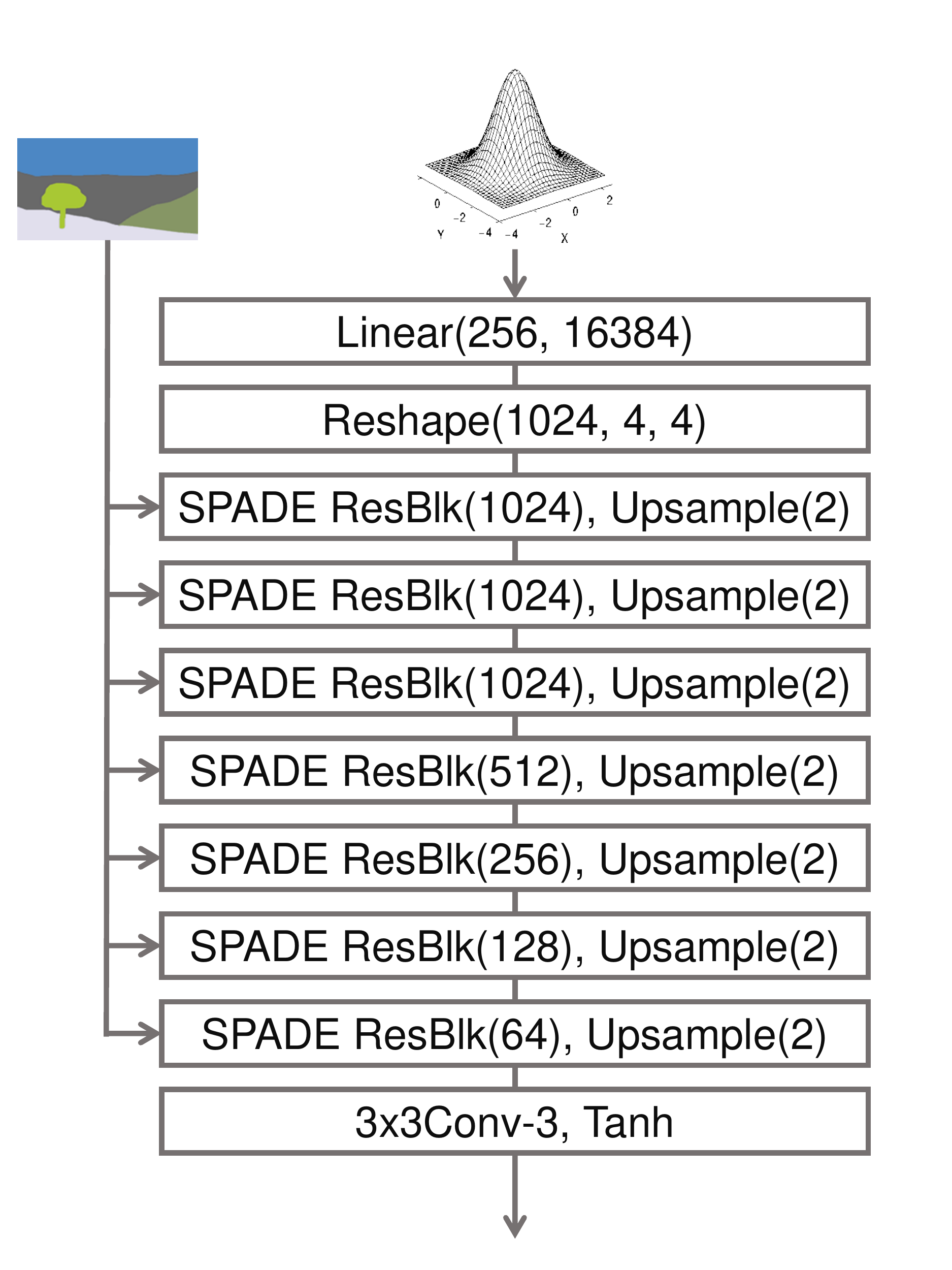}
 \vspace{-2mm}
  \caption{SPADE Generator. Different from prior image generators~\cite{isola2017image,wang2018pix2pixHD}, the semantic segmentation mask is passed to the generator through the proposed SPADE ResBlks in Figure~\ref{fig::supp_spade_resblk_arch}.}
  \label{fig::supp_gen}
  \vspace{-2mm}
\end{figure}

\mysubsection{Generator.} The architecture of the generator consists of a series of the proposed SPADE ResBlks with nearest neighbor upsampling. We train our network using 8 GPUs simultaneously and use the synchronized version of the BatchNorm. We apply the Spectral Norm~\cite{miyato2018spectral} to all the convolutional layers in the generator. The architectures of the proposed SPADE and SPADE ResBlk are given in Figure~\ref{fig::supp_spade_arch} and Figure~\ref{fig::supp_spade_resblk_arch}, respectively. The architecture of the generator is shown in Figure~\ref{fig::supp_gen}.

\begin{figure}[!t]
	\centering
\includegraphics[trim=0.00in 0.0in 0in 0in, width=0.3\textwidth]{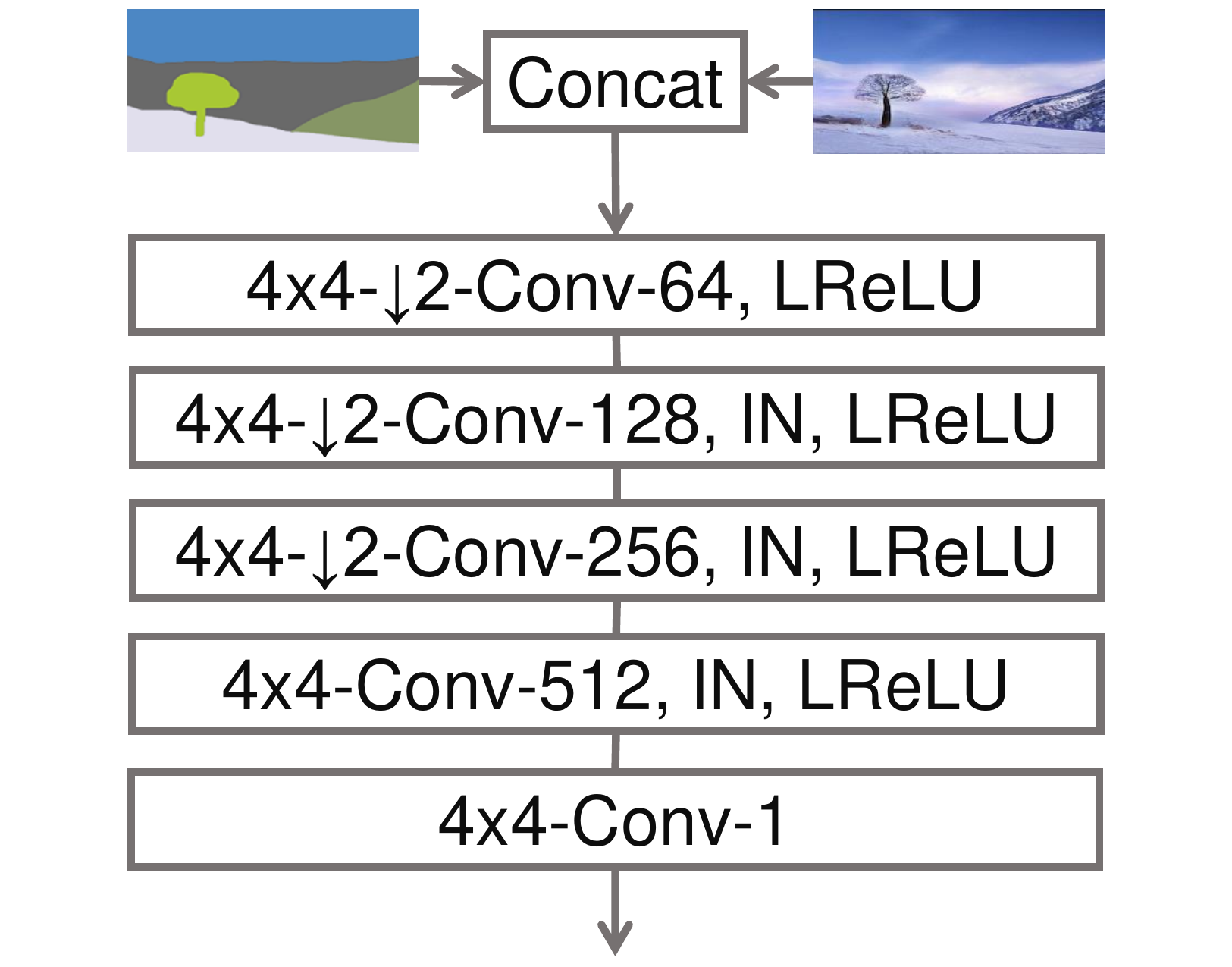}
 \vspace{-2mm}
  \caption{Our discriminator design largely follows that in the \pixtopixhd~\cite{wang2018pix2pixHD}. It takes the concatenation the segmentation map and the image as input. It is based on the PatchGAN~\cite{isola2017image}. Hence, the last layer of the discriminator is a convolutional layer.}
 \label{fig::supp_dis}
 \vspace{-2mm} 
\end{figure}

\mysubsection{Discriminator.} The architecture of the discriminator follows the one used in the \pixtopixhd method~\cite{wang2018pix2pixHD}, which uses a multi-scale design with the InstanceNorm (IN). The only difference is that we apply the Spectral Norm to all the convolutional layers of the discriminator. The details of the discriminator architecture is shown in Figure~\ref{fig::supp_dis}.

\begin{figure}[!t]
\centering
\includegraphics[trim=0.00in 0.0in 0in 0in, width=0.3\textwidth]{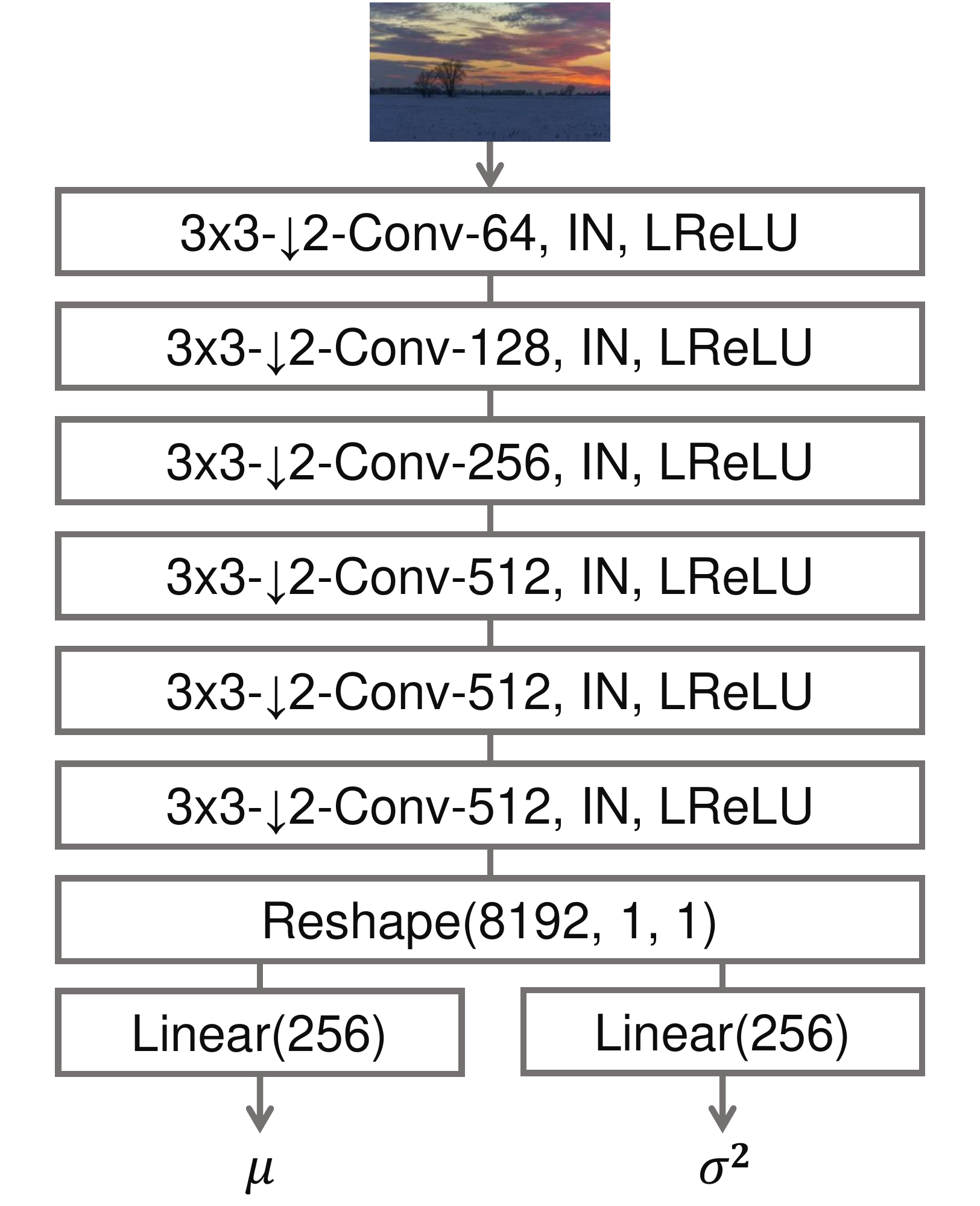}
  \vspace{-2mm}
  \caption{The image encoder consists a series of convolutional layers with stride 2 followed by two linear layers that output a mean vector $\mu$ and a variance vector $\sigma$.} \label{fig::supp_encoder}
  \vspace{-2mm}
\end{figure}

\mysubsection{Image Encoder.} The image encoder consists of 6 stride-2 convolutional layers followed by two linear layers to produce the mean and variance of the output distribution as shown in Figure~\ref{fig::supp_encoder}.

\mysubsection{Learning objective.} We use the learning objective function in the \pixtopixhd work~\cite{wang2018pix2pixHD} except that we replace its LSGAN loss~\cite{mao2017least} term with the Hinge loss term~\cite{lim2017geometric,miyato2018spectral,zhang2018self}. We use the same weighting among the loss terms in the objective function as that in the \pixtopixhd work.

\texttt{}
\begin{figure}[!bh]
\centering
\includegraphics[trim=0.00in 0.0in 0in 0in, width=0.3\textwidth]{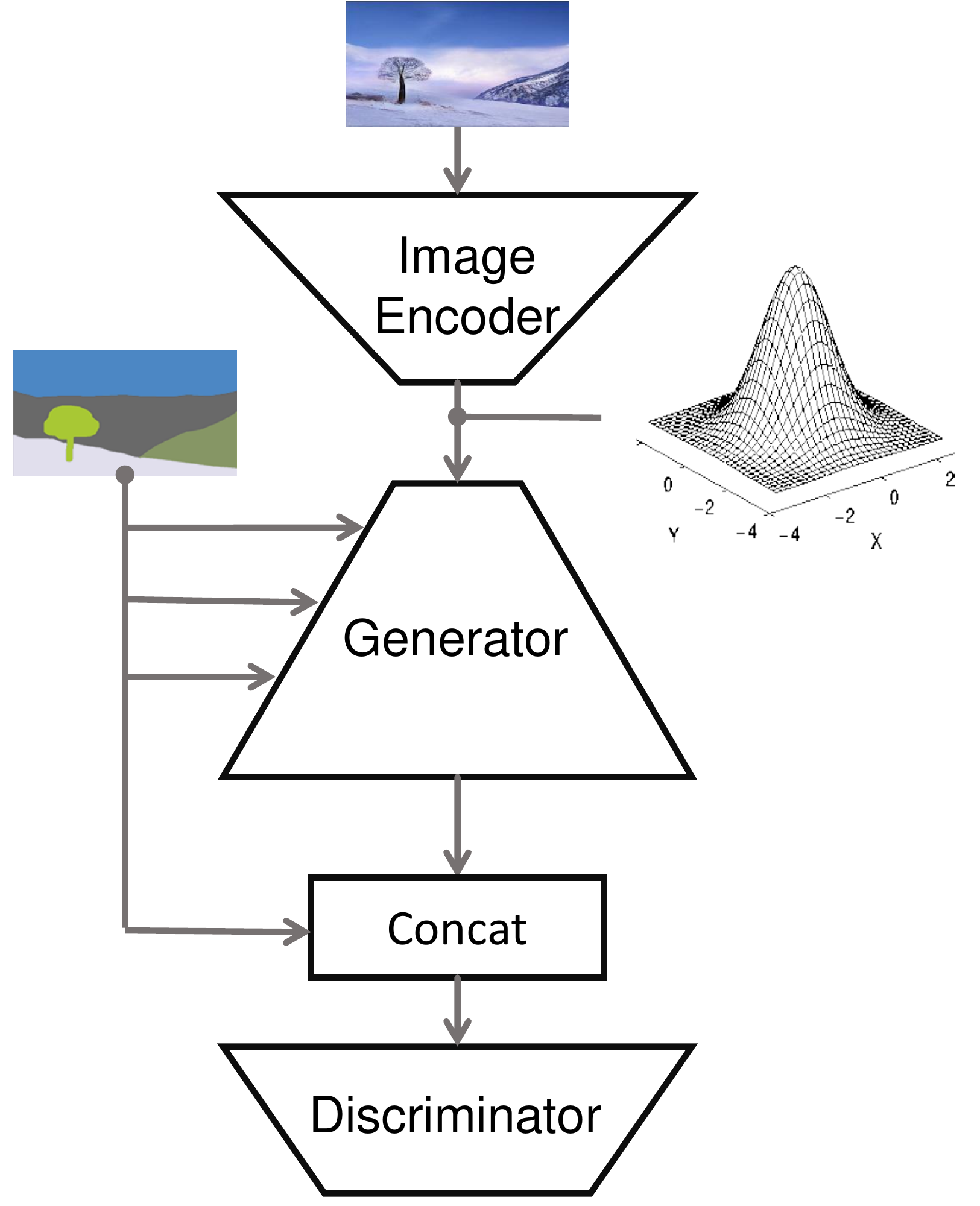}
 \vspace{-2mm} 
  \caption{The image encoder encodes a real image to a latent representation for generating a mean vector and a variance vector. They are used to compute the noise input to the generator via the reparameterization trick~\cite{kingma2013auto}. The generator also takes the segmentation mask of the input image as input via the proposed SPADE ResBlks. The discriminator takes concatenation of the segmentation mask and the output image from the generator as input and aims to classify that as fake.}
 \label{fig::supp_overview}
 \vspace{-2mm}
\end{figure}

When training the proposed framework with the image encoder for multi-modal synthesis and style-guided image synthesis, we include a KL Divergence loss:
\begin{equation}
    \mathcal{L}_{\text{KLD}} = \mathcal{D}_{\text{KL}}( q(\mathbf{z}|\mathbf{x}) || p(\mathbf{z})) \nonumber
\end{equation}
where the prior distribution $p(\mathbf{z})$ is a standard Gaussian distribution and the variational distribution $q$ is fully determined by a mean vector and a variance vector~\cite{kingma2013auto}. We use the reparamterization trick~\cite{kingma2013auto} for back-propagating the gradient from the generator to the image encoder. The weight for the KL Divergence loss is $0.05$. 

In Figure~\ref{fig::supp_overview}, we overview the training data flow. The image encoder encodes a real image to a mean vector and a variance vector. They are used to compute the noise input to the generator via the reparameterization trick~\cite{kingma2013auto}. The generator also takes the segmentation mask of the input image as input with the proposed SPADE ResBlks. The discriminator takes concatenation of the segmentation mask and the output image from the generator as input and aims to classify that as fake.

\mysubsection{Training details.}  We perform $200$ epochs of training on the Cityscapes and ADE20K datasets, $100$ epochs of training on the COCO-Stuff dataset, and $50$ epochs of training on the Flickr Landscapes dataset. The image sizes are $256\times 256$, except the Cityscapes at $512\times 256$. We linearly decay the learning rate to $0$ from epoch $100$ to $200$ for the Cityscapes and ADE20K datasets. The batch size is $32$. We initialize the network weights using thes Glorot initialization~\cite{glorot2010understanding}.

\clearpage
\section{Additional Ablation Study}

\bgroup
\def\arraystretch{1.1}
\addtolength{\tabcolsep}{-2.5pt}  
\begin{table}[!h]
	\centering
	\small
    \begin{tabular}{l||c|c|c}
	    Method & COCO. & ADE. & City.\\
		\hline
Ours & 35.2 & 38.5 & 62.3 \\
Ours w/o  Perceptual loss & 24.7 & 30.1 & 57.4\\   
Ours w/o GAN feature matching loss & 33.2 & 38.0 & 62.2\\   
Ours w/\ \ \ a deeper discriminator & 34.9 & 38.3 & 60.9\\
\hline
\pixtopixplusplus w/\ \ \  SPADE & 
                34.4 & 39.0 & 62.2 \\
\pixtopixplusplus &
                32.7 & 38.3 & 58.8\\
\pixtopixplusplus w/o Sync BatchNorm & 
                27.4 & 31.8 & 51.1\\
\pixtopixplusplus w/o Sync BatchNorm, & 
                26.0 & 31.9 & 52.3\\
\quad \quad \quad \medspace and w/o Spectral Norm & & &\\
\hline
\pixtopixhd ~\cite{wang2018pix2pixHD}& 14.6 & 20.3 & 58.3\\
	\end{tabular}
	\caption{Additional ablation study results using the mIoU metric: the table shows that both the perceptual loss and GAN feature matching loss terms are important. Making the discriminator deeper does not lead to a performance boost. The table also shows that all the components (Synchronized BatchNorm, Spectral Norm, TTUR, the Hinge loss objective, and the SPADE) used in the proposed method helps our strong baseline, \pixtopixplusplus.}
	\label{tbl::moreablation}		
 \end{table}
 \addtolength{\tabcolsep}{2.5pt}  
 \egroup

Table~\ref{tbl::moreablation} provides additional ablation study results analyzing the contribution of individual components in the proposed method. We first find that both of the perceptual loss and GAN feature matching loss inherited from the learning objective function of the \pixtopixhd~\cite{wang2018pix2pixHD} are important. Removing any of them leads to a performance drop. We also find that increasing the depth of the discriminator by inserting one more convolutional layer to the top of the \pixtopixhd discriminator does not improve the results.

In Table~\ref{tbl::moreablation}, we also analyze the effectiveness of each component used in our strong baseline, the \pixtopixplusplus method, derived from the \pixtopixhd method. We found that the Spectral Norm, synchronized BatchNorm, TTUR~\cite{heusel2017gans}, and the hinge loss objective all contribute to the performance boost. Adding the SPADE to the strong baseline further improves the performance. Note that the \pixtopixplusplus w/o Sync BatchNorm and w/o Spectral Norm still differs from the \pixtopixhd in that it uses the hinge loss objective, TTUR, a large batch size, and the Glorot initialization~\cite{glorot2010understanding}.

\section{Additional Results}

In Figure~\ref{fig::supp_coco},~\ref{fig::supp_coco2}, and~\ref{fig::supp_ade}, we show additional synthesis results from the proposed method on the COCO-Stuff and ADE20K datasets with comparisons to those from the \crn~\cite{chen2017photographic} and \pixtopixhd~\cite{wang2018pix2pixHD} methods.

In Figure~\ref{fig::supp_adeoutdoor} and~\ref{fig::supp_cityscapes}, we show additional synthesis results from the proposed method on the ADE20K-outdoor and Cityscapes datasets with comparison to those from the \crn~\cite{chen2017photographic}, \sims~\cite{qi2018semi}, and \pixtopixhd~\cite{wang2018pix2pixHD} methods. 

In Figure~\ref{fig::supp_flickr}, we show additional multi-modal synthesis results from the proposed method. As sampling different $\mathbf{z}$ from a standard multivariate Gaussian distribution, we synthesize images of diverse appearances.

In the accompanying \href{https://github.com/NVlabs/SPADE}{video}, we demonstrate our semantic image synthesis interface. We show how a user can create photorealistic landscape images by painting semantic labels on a canvas. We also show how a user can synthesize images of diverse appearances for the same semantic segmentation mask as well as transfer the appearance of a provided style image to the synthesized one.

\addtolength{\tabcolsep}{-4.5pt}    
\newlength{\mywidthsuppcoco}
\setlength{\mywidthsuppcoco}{0.193\textwidth}
\setlength{\fboxsep}{0pt}%
\setlength{\fboxrule}{0.5pt}%
\newcommand{\insertrow}[1]{
\fbox{\includegraphics[width=\mywidthsuppcoco, height=1.0in]{{"supplementary/coco/label/#1"}.jpg}} & 
\fbox{\includegraphics[width=\mywidthsuppcoco, height=1.0in]{{"supplementary/coco/gt/#1"}.jpg}} & \fbox{\includegraphics[width=\mywidthsuppcoco, height=1.0in]{{"supplementary/coco/crn/#1"}.jpg}} &
\fbox{\includegraphics[width=\mywidthsuppcoco, height=1.0in]{{"supplementary/coco/pix2pixhd/#1"}.jpg}} &
\fbox{\includegraphics[width=\mywidthsuppcoco, height=1.0in]{{"supplementary/coco/ours/#1"}.jpg}} \\
}
\bgroup
\def\arraystretch{1}
\begin{figure*}[t!]
\begin{tabular}{ccccc}
Label & Ground Truth & \crn & \pixtopixhd & \textbf{Ours} \\
\insertrow{000000134689}
\insertrow{000000193245}
\insertrow{000000024027}
\insertrow{000000051314}
\insertrow{000000445602}
\insertrow{000000407650}
\insertrow{000000436551}
\insertrow{000000383921}

\end{tabular}
	\caption{Additional results with comparison to those from the \crn~\cite{chen2017photographic} and \pixtopixhd~\cite{wang2018pix2pixHD} methods on the COCO-Stuff dataset.}\label{fig::supp_coco}
\end{figure*}
 \egroup
 \addtolength{\tabcolsep}{4.5pt}
\vfill

\addtolength{\tabcolsep}{-4.5pt}    
\setlength{\mywidthsuppcoco}{0.193\textwidth}
\setlength{\fboxsep}{0pt}
\setlength{\fboxrule}{0.5pt}
\renewcommand{\insertrow}[1]{
\fbox{\includegraphics[width=\mywidthsuppcoco, height=1.0in]{{"supplementary/coco/label/#1"}.jpg}} & 
\fbox{\includegraphics[width=\mywidthsuppcoco, height=1.0in]{{"supplementary/coco/gt/#1"}.jpg}} & \fbox{\includegraphics[width=\mywidthsuppcoco, height=1.0in]{{"supplementary/coco/crn/#1"}.jpg}} &
\fbox{\includegraphics[width=\mywidthsuppcoco, height=1.0in]{{"supplementary/coco/pix2pixhd/#1"}.jpg}} &
\fbox{\includegraphics[width=\mywidthsuppcoco, height=1.0in]{{"supplementary/coco/ours/#1"}.jpg}} \\
}
\bgroup
\def\arraystretch{1}
\begin{figure*}[t!]
\begin{tabular}{ccccc}
Label & Ground Truth & \crn & \pixtopixhd & \textbf{Ours} \\
\insertrow{000000250758}
\insertrow{000000490470}
\insertrow{000000139684}
\insertrow{000000062025}
\insertrow{000000074733}
\insertrow{000000030785}
\insertrow{000000206994}
\insertrow{000000178744}
\end{tabular}
	\caption{Additional results with comparison to those from the \crn~\cite{chen2017photographic} and \pixtopixhd~\cite{wang2018pix2pixHD} methods on the COCO-Stuff dataset.} \label{fig::supp_coco2}	
	\end{figure*}
 \egroup
 \addtolength{\tabcolsep}{4.5pt}
\vfill

\addtolength{\tabcolsep}{-4.5pt}    
\newlength{\mywidthsuppade}
\newlength{\myheightsuppade}
\setlength{\mywidthsuppade}{0.192\textwidth}
\setlength{\myheightsuppade}{1.1in}
\setlength{\fboxsep}{0pt}%
\setlength{\fboxrule}{0.5pt}%
\renewcommand{\insertrow}[1]{
\fbox{\includegraphics[width=\mywidthsuppade, height=\myheightsuppade]{{"supplementary/ade20k/label/ADE_val_#1"}.jpg}} & 
\fbox{\includegraphics[width=\mywidthsuppade, height=\myheightsuppade]{{"supplementary/ade20k/gt/ADE_val_#1"}.jpg}} & \fbox{\includegraphics[width=\mywidthsuppade, height=\myheightsuppade]{{"supplementary/ade20k/crn/ADE_val_#1"}.jpg}} &
\fbox{\includegraphics[width=\mywidthsuppade, height=\myheightsuppade]{{"supplementary/ade20k/pix2pixhd/ADE_val_#1"}.jpg}} &
\fbox{\includegraphics[width=\mywidthsuppade, height=\myheightsuppade]{{"supplementary/ade20k/ours/ADE_val_#1"}.jpg}} \\
}
\bgroup
\def\arraystretch{1}
\begin{figure*}[t!]
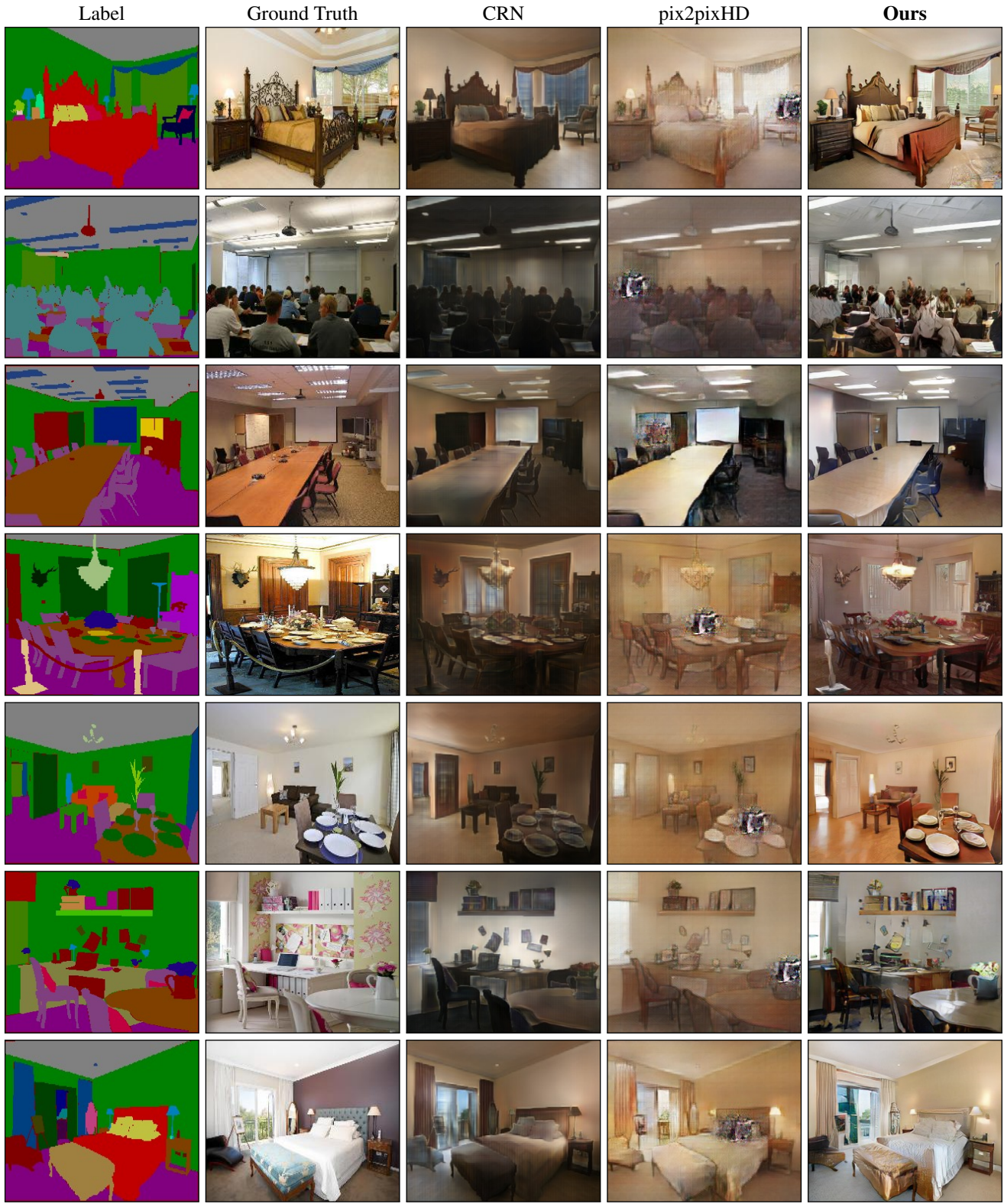

\begin{tabular}{ccccc}
Label & Ground Truth & CRN & \pixtopixhd & \textbf{Ours} \\
\insertrow{00000135}
\insertrow{00000251}
\insertrow{00000268}
\insertrow{00000309}
\insertrow{00000318}
\insertrow{00001012}
\insertrow{00001136}
\end{tabular}
	\caption{Additional results with comparison to those from the \crn~\cite{chen2017photographic} and \pixtopixhd~\cite{wang2018pix2pixHD} methods on the ADE20K dataset.} \label{fig::supp_ade}
 \end{figure*}
 \egroup
 \addtolength{\tabcolsep}{4.5pt}
 \vfill

\addtolength{\tabcolsep}{-4.5pt}
\newlength{\mywidthsuppadeoutdoor}
\newlength{\myheightsuppadeoutdoor}
\setlength{\mywidthsuppadeoutdoor}{0.16\textwidth}
\setlength{\myheightsuppadeoutdoor}{0.73in}
\setlength{\fboxsep}{0pt}%
\setlength{\fboxrule}{0.5pt}%
\renewcommand{\insertrow}[1]{
\fbox{\includegraphics[width=\mywidthsuppadeoutdoor, height=\myheightsuppadeoutdoor]{{"supplementary/ade20k/label/ADE_val_#1"}.jpg}} &
\fbox{\includegraphics[width=\mywidthsuppadeoutdoor, height=\myheightsuppadeoutdoor]{{"supplementary/ade20k/gt/ADE_val_#1"}.jpg}} & \fbox{\includegraphics[width=\mywidthsuppadeoutdoor, height=\myheightsuppadeoutdoor]{{"supplementary/ade20k/crn/ADE_val_#1"}.jpg}} &
\fbox{\includegraphics[width=\mywidthsuppadeoutdoor, height=\myheightsuppadeoutdoor]{{"supplementary/ade20k/sims/ADE_val_#1"}.jpg}} &
\fbox{\includegraphics[width=\mywidthsuppadeoutdoor, height=\myheightsuppadeoutdoor]{{"supplementary/ade20k/pix2pixhd/ADE_val_#1"}.jpg}} &
\fbox{\includegraphics[width=\mywidthsuppadeoutdoor, height=\myheightsuppadeoutdoor]{{"supplementary/ade20k/ours/ADE_val_#1"}.jpg}} \\
}
\bgroup
\def\arraystretch{1}
\begin{figure*}[t!]
\begin{tabular}{cccccc}
Label & Ground Truth & \crn & \sims & \pixtopixhd & \textbf{Ours} \\
\insertrow{00000262}
\insertrow{00000289}
\insertrow{00000291}
\insertrow{00000389}
\insertrow{00000574}
\insertrow{00000733}
\insertrow{00001112}
\insertrow{00001344}
\insertrow{00001390}
\insertrow{00001582}
\insertrow{00001795}
\end{tabular}
	\caption{Additional results with comparison to those from the \crn~\cite{chen2017photographic}, \sims~\cite{qi2018semi}, and \pixtopixhd~\cite{wang2018pix2pixHD} methods on the ADE20K-outdoor dataset.} \label{fig::supp_adeoutdoor}		
 \end{figure*}
 \egroup
 \addtolength{\tabcolsep}{4.5pt}
\vfill

\addtolength{\tabcolsep}{-4.5pt}    
\newlength{\mywidthsuppcityscapes}
\setlength{\mywidthsuppcityscapes}{0.327\textwidth}
\setlength{\fboxsep}{0pt}%
\setlength{\fboxrule}{0.5pt}%
\renewcommand{\insertrow}[1]{
Label & Ground Truth & \textbf{Ours} \\ 
\fbox{\includegraphics[width=\mywidthsuppcityscapes]{{"supplementary/cityscapes/label/#1_gtFine_color"}.jpg}} & 
\fbox{\includegraphics[width=\mywidthsuppcityscapes]{{"supplementary/cityscapes/gt/#1_leftImg8bit"}.jpg}} & \fbox{\includegraphics[width=\mywidthsuppcityscapes]{{"supplementary/cityscapes/ours/#1_leftImg8bit"}.jpg}} \\
\fbox{\includegraphics[width=\mywidthsuppcityscapes]{{"supplementary/cityscapes/crn/#1_leftImg8bit"}.jpg}} &
\fbox{\includegraphics[width=\mywidthsuppcityscapes]{{"supplementary/cityscapes/sims/#1_leftImg8bit"}.jpg}} &
\fbox{\includegraphics[width=\mywidthsuppcityscapes]{{"supplementary/cityscapes/pix2pixhd/#1_leftImg8bit"}.jpg}} \\
CRN & SIMS & \pixtopixhd \\
}
\bgroup
\def\arraystretch{0.5}
\begin{figure*}[t!]
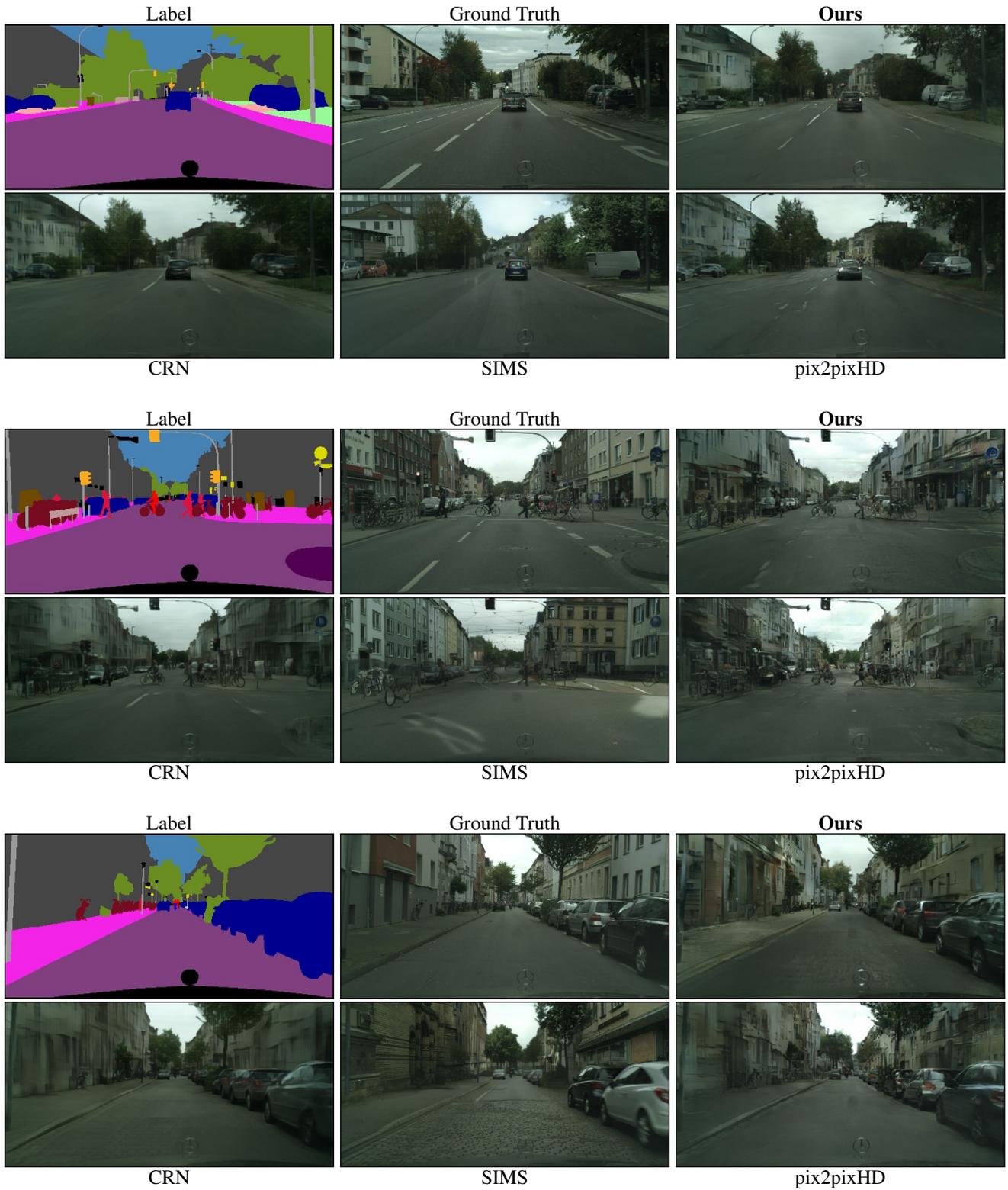

\begin{tabular}{ccc}
\insertrow{lindau_000004_000019}
\addlinespace[16pt]
\insertrow{munster_000013_000019}
\addlinespace[16pt]
\insertrow{munster_000096_000019}
\end{tabular}
	\caption{Additional results with comparison to those from the \crn~\cite{chen2017photographic}, \sims~\cite{qi2018semi}, and \pixtopixhd~\cite{wang2018pix2pixHD} methods on the Cityscapes dataset.} \label{fig::supp_cityscapes}		
 \end{figure*}
 \egroup
 \addtolength{\tabcolsep}{4.5pt}
\vfill

\addtolength{\tabcolsep}{-4.5pt}    
\newlength{\mywidthsuppflickr}
\setlength{\mywidthsuppflickr}{0.19\textwidth}
\setlength{\fboxsep}{0pt}%
\setlength{\fboxrule}{0.5pt}%
\bgroup
\def\arraystretch{1}
\begin{figure*}[t!]
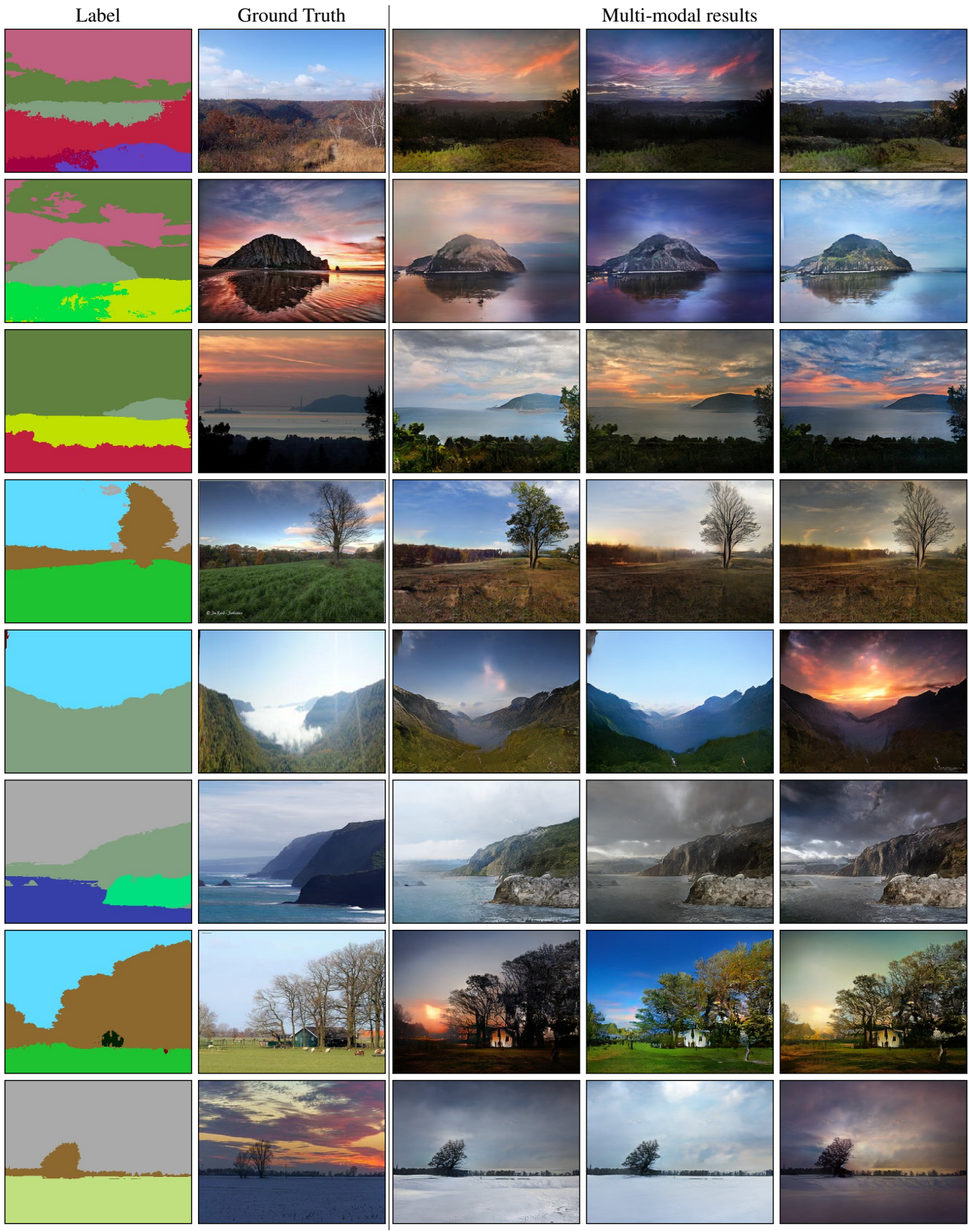

	\begin{tabular}{cc|ccc}
		Label & Ground Truth & & Multi-modal results & \\
		\fbox{\includegraphics[width=\mywidthsuppflickr, height=1.0in]{{"supplementary/flickr/label/2011-11-03_19_08_54"}.jpg}} & 
		\fbox{\includegraphics[width=\mywidthsuppflickr, height=1.0in]{{"supplementary/flickr/gt/2011-11-03_19_08_54"}.jpg}} & \fbox{\includegraphics[width=\mywidthsuppflickr, height=1.0in]{{"supplementary/flickr/ours/2011-11-03_19_08_54_01"}.jpg}} &
		\fbox{\includegraphics[width=\mywidthsuppflickr, height=1.0in]{{"supplementary/flickr/ours/2011-11-03_19_08_54_02"}.jpg}} &
		\fbox{\includegraphics[width=\mywidthsuppflickr, height=1.0in]{{"supplementary/flickr/ours/2011-11-03_19_08_54_03"}.jpg}} \\ 
		
		\fbox{\includegraphics[width=\mywidthsuppflickr, height=1.0in]{{"supplementary/flickr/label/2013-01-20_10_29_23"}.jpg}} & 
		\fbox{\includegraphics[width=\mywidthsuppflickr, height=1.0in]{{"supplementary/flickr/gt/2013-01-20_10_29_23"}.jpg}} & \fbox{\includegraphics[width=\mywidthsuppflickr, height=1.0in]{{"supplementary/flickr/ours/2013-01-20_10_29_23_01"}.jpg}} &
		\fbox{\includegraphics[width=\mywidthsuppflickr, height=1.0in]{{"supplementary/flickr/ours/2013-01-20_10_29_23_04"}.jpg}} &
		\fbox{\includegraphics[width=\mywidthsuppflickr, height=1.0in]{{"supplementary/flickr/ours/2013-01-20_10_29_23_05"}.jpg}} \\
		
		\fbox{\includegraphics[width=\mywidthsuppflickr, height=1.0in]{{"supplementary/flickr/label/2012-02-04_22_19_29"}.jpg}} & 
		\fbox{\includegraphics[width=\mywidthsuppflickr, height=1.0in]{{"supplementary/flickr/gt/2012-02-04_22_19_29"}.jpg}} & \fbox{\includegraphics[width=\mywidthsuppflickr, height=1.0in]{{"supplementary/flickr/ours/2012-02-04_22_19_29_01"}.jpg}} &
		\fbox{\includegraphics[width=\mywidthsuppflickr, height=1.0in]{{"supplementary/flickr/ours/2012-02-04_22_19_29_02"}.jpg}} &
		\fbox{\includegraphics[width=\mywidthsuppflickr, height=1.0in]{{"supplementary/flickr/ours/2012-02-04_22_19_29_03"}.jpg}} \\
		
		\fbox{\includegraphics[width=\mywidthsuppflickr, height=1.0in]{{"supplementary/flickr/label/2013-11-11_11_47_53"}.jpg}} & 
		\fbox{\includegraphics[width=\mywidthsuppflickr, height=1.0in]{{"supplementary/flickr/gt/2013-11-11_11_47_53"}.jpg}} & \fbox{\includegraphics[width=\mywidthsuppflickr, height=1.0in]{{"supplementary/flickr/ours/2013-11-11_11_47_53_01"}.jpg}} &
		\fbox{\includegraphics[width=\mywidthsuppflickr, height=1.0in]{{"supplementary/flickr/ours/2013-11-11_11_47_53_02"}.jpg}} &
		\fbox{\includegraphics[width=\mywidthsuppflickr, height=1.0in]{{"supplementary/flickr/ours/2013-11-11_11_47_53_05"}.jpg}} \\
		
		\fbox{\includegraphics[width=\mywidthsuppflickr, height=1.0in]{{"supplementary/flickr/label/2014-12-02_11_30_28"}.jpg}} & 
		\fbox{\includegraphics[width=\mywidthsuppflickr, height=1.0in]{{"supplementary/flickr/gt/2014-12-02_11_30_28"}.jpg}} & \fbox{\includegraphics[width=\mywidthsuppflickr, height=1.0in]{{"supplementary/flickr/ours/2014-12-02_11_30_28_00"}.jpg}} &
		\fbox{\includegraphics[width=\mywidthsuppflickr, height=1.0in]{{"supplementary/flickr/ours/2014-12-02_11_30_28_02"}.jpg}} &
		\fbox{\includegraphics[width=\mywidthsuppflickr, height=1.0in]{{"supplementary/flickr/ours/2014-12-02_11_30_28_05"}.jpg}} \\
		
		\fbox{\includegraphics[width=\mywidthsuppflickr, height=1.0in]{{"supplementary/flickr/label/2016-02-28_18_55_15"}.jpg}} & 
		\fbox{\includegraphics[width=\mywidthsuppflickr, height=1.0in]{{"supplementary/flickr/gt/2016-02-28_18_55_15"}.jpg}} & \fbox{\includegraphics[width=\mywidthsuppflickr, height=1.0in]{{"supplementary/flickr/ours/2016-02-28_18_55_15_00"}.jpg}} &
		\fbox{\includegraphics[width=\mywidthsuppflickr, height=1.0in]{{"supplementary/flickr/ours/2016-02-28_18_55_15_01"}.jpg}} &
		\fbox{\includegraphics[width=\mywidthsuppflickr, height=1.0in]{{"supplementary/flickr/ours/2016-02-28_18_55_15_02"}.jpg}} \\
		
		\fbox{\includegraphics[width=\mywidthsuppflickr, height=1.0in]{{"supplementary/flickr/label/2016-06-15_05_44_04"}.jpg}} & 
		\fbox{\includegraphics[width=\mywidthsuppflickr, height=1.0in]{{"supplementary/flickr/gt/2016-06-15_05_44_04"}.jpg}} & \fbox{\includegraphics[width=\mywidthsuppflickr, height=1.0in]{{"supplementary/flickr/ours/2016-06-15_05_44_04_00"}.jpg}} &
		\fbox{\includegraphics[width=\mywidthsuppflickr, height=1.0in]{{"supplementary/flickr/ours/2016-06-15_05_44_04_01"}.jpg}} &
		\fbox{\includegraphics[width=\mywidthsuppflickr, height=1.0in]{{"supplementary/flickr/ours/2016-06-15_05_44_04_03"}.jpg}} \\
		
		\fbox{\includegraphics[width=\mywidthsuppflickr, height=1.0in]{{"supplementary/flickr/label/2016-12-26_08_29_26"}.jpg}} & 
		\fbox{\includegraphics[width=\mywidthsuppflickr, height=1.0in]{{"supplementary/flickr/gt/2016-12-26_08_29_26"}.jpg}} & \fbox{\includegraphics[width=\mywidthsuppflickr, height=1.0in]{{"supplementary/flickr/ours/2016-12-26_08_29_26_00"}.jpg}} &
		\fbox{\includegraphics[width=\mywidthsuppflickr, height=1.0in]{{"supplementary/flickr/ours/2016-12-26_08_29_26_04"}.jpg}} &
		\fbox{\includegraphics[width=\mywidthsuppflickr, height=1.0in]{{"supplementary/flickr/ours/2016-12-26_08_29_26_03"}.jpg}} \\
	\end{tabular}
	\caption{Additional multi-modal synthesis results on the Flickr Landscapes Dataset. By sampling latent vectors from a standard Gaussian distribution, we synthesize images of diverse appearances.} \label{fig::supp_flickr}		
\end{figure*}
\egroup
\addtolength{\tabcolsep}{4.5pt}
\vfill

\end{document}